\documentclass[manuscript,screen,nonacm]{acmart}

\AtBeginDocument{%
  \providecommand\BibTeX{{%
    \normalfont B\kern-0.5em{\scshape i\kern-0.25em b}\kern-0.8em\TeX}}}

\setcopyright{none}
\renewcommand\footnotetextcopyrightpermission[1]{}
\usepackage{graphicx}
\usepackage{subcaption}
\usepackage{bm}
\usepackage{graphicx}
\usepackage{placeins}
\usepackage{multirow}
\usepackage{wasysym}

\newcommand{\sX}{\mathcal{X}}
\newcommand{\sY}{\mathcal{Y}}

\newcommand{\upweightfactor}{\lambda_{\text{up}}}
\newcommand{\bx}{\bm{x}}

 \newcommand{\indep}{\perp\!\!\!\!\perp} 

\begin{document}

\title[Measuring the Fairness of Explanations]{The Road to Explainability is Paved with Bias: Measuring the Fairness of Explanations}
\author{Aparna Balagopalan
}
\affiliation{\institution{Massachusetts Institute of Technology}
\country{USA}}
\email{aparnab@mit.edu}
\author{Haoran Zhang}
\affiliation{\institution{Massachusetts Institute of Technology}
\country{USA}}
\email{haoranz@mit.edu}
\author{Kimia Hamidieh}
\affiliation{
\institution{University of Toronto},
\institution{Vector Institute}
\country{Canada}}
\email{kimia@cs.toronto.edu}
\author{Thomas Hartvigsen}
\affiliation{\institution{Massachusetts Institute of Technology}
\country{USA}}
\email{tomh@mit.edu}
\author{Frank Rudzicz}
\affiliation{
\institution{University of Toronto}, \institution{Vector Institute}, 
\institution{Unity Health Toronto}
\country{Canada}}
\email{frank@cs.toronto.edu}
\author{Marzyeh Ghassemi}
\affiliation{\institution{Massachusetts Institute of Technology},
\institution{Vector Institute}
\country{USA}}
\email{mghassem@mit.edu}

\renewcommand{\shortauthors}{Balagopalan, Zhang, Hamidieh, Hartvigsen, Rudzicz, Ghassemi}

\begin{abstract}
Machine learning models in safety-critical settings like healthcare are often ``blackboxes'': they contain a large number of parameters which are not transparent to users. Post-hoc explainability methods where a simple, human-interpretable model imitates the behavior of these blackbox models are often proposed to help users trust model predictions. In this work, we audit the quality of such explanations for different protected subgroups using real data from four settings in finance, healthcare, college admissions, and the US justice system. Across two different blackbox model architectures and four popular explainability methods, we find that the approximation quality of explanation models, also known as the \emph{fidelity}, differs significantly between subgroups.  
We also demonstrate that pairing explainability methods with recent advances in robust machine learning can improve explanation fairness in some settings. However, we highlight the importance of communicating details of non-zero fidelity gaps to users, since a single solution might not exist across all settings. Finally, we discuss the implications of unfair explanation models as a 
challenging and understudied problem facing the machine learning community.

\end{abstract}



\maketitle

\section{Introduction}
Machine learning (ML) models are increasingly used in safety-critical settings like healthcare~\cite{chen2020ethical, tuggener2019automated}, college admissions~\cite{jamison2017applying}, and law~\cite{alarie2016using}.
Several studies have shown that human decisions can become more accurate when assisted by such ML models \cite{buccinca2020proxy,tschandl2020human,wang2020human}. However, many ML models are ``blackboxes''---they might have too many parameters or be proprietary---and cannot explain their predictions in ways humans understand~\cite{rudin2019stop}.
In such scenarios, users may struggle to understand a model's outputs enough to trust and use its predictions~\cite{dietvorst2015algorithm,doshi2017towards,lipton2018mythos}.

\begin{figure}
\centering
\includegraphics[width=0.9\textwidth]{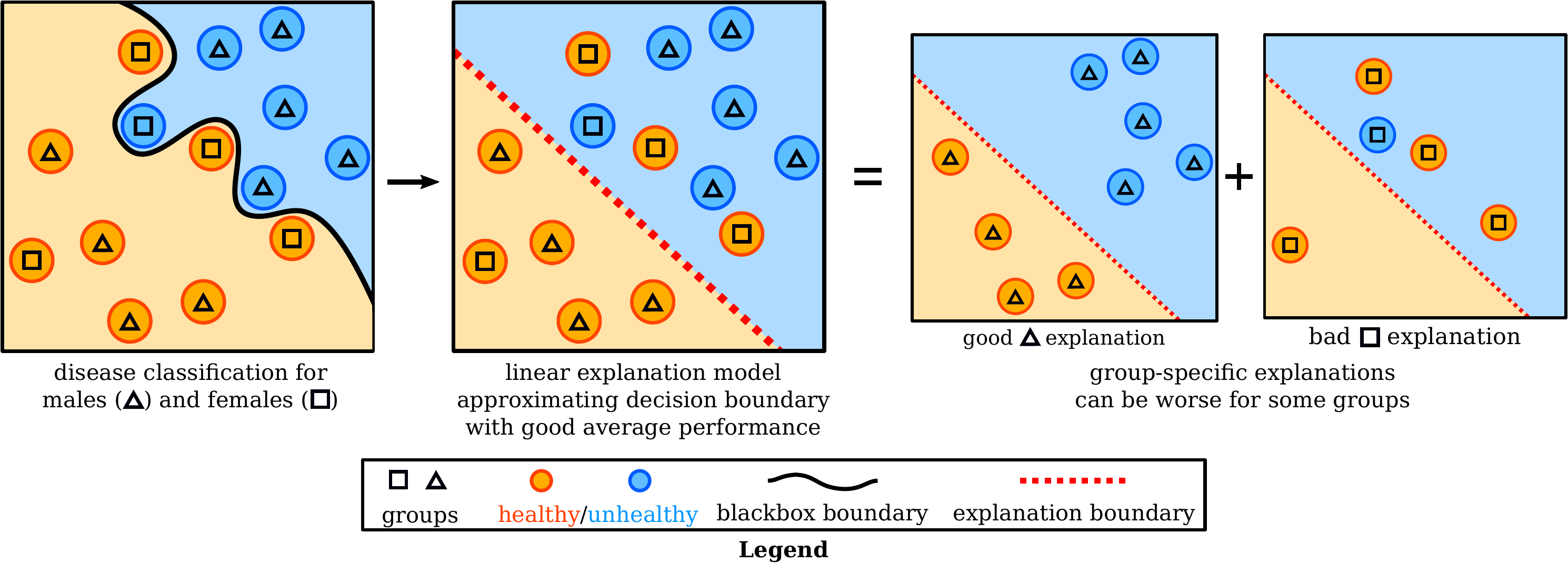}
\caption{An example of an unfair global explanation model. Orange and Blue circles indicate predicted classes, \textit{healthy} or \textit{unhealthy}, respectively. $\square$ and $\triangle$ denote group membership. The red dashed line is a linear explainability model fit to approximate the black blackbox decision boundary. The two figures on the right show that the linear approximate is worse for the $\square$ group.}\label{fig:local_global_expls}
\end{figure}

\emph{Post-hoc explainability} methods have recently begun helping users better understand why blackbox models make certain predictions \cite{ribeiro2016model,ribeiro2018anchors,lakkaraju2019faithful}. A popular post-hoc approach is to train simple, human-interpretable models to \textit{imitate} a blackbox model's behaviour \cite{ribeiro2016should} by maximizing the congruity between simple approximations and blackbox model predictions. Such approximation quality is known as \textit{fidelity}~\cite{craven1995extracting}. 
Then, the simpler model can be used either as a new stand-alone model or to explain one prediction at a time~\cite{tan2018distill, ribeiro2016should}. By highlighting important inputs, these explainability methods provide a path towards helping users trust machine learning models in high-impact settings \cite{ribeiro2016model}.

However, it remains unknown when and if these models approximate behavior \textit{fairly}.
If fidelity differs between different pre-defined groups (e.g., between demographics) in a dataset, explainability methods may perpetuate machine bias by encouraging users to trust model predictions for some people but not others.
In this work, we study \textit{to what degree do gaps in fidelity exist between subgroups?}

To answer this question, we measure group-wise fidelity for different post-hoc explanation methods on real tabular datasets that include group membership. 
Intuitively, an explanation model is \emph{fair} if it has equally high fidelity for all protected groups. This definition is similar to common group fairness definitions which seek to eliminate gaps in predictive performance across groups~\cite{pleiss2017,hardt2016equalityofopportunity,zhang2022improving}. We introduce two definitions of \textit{fidelity gaps}, or disparities in fidelity across different subgroups. Using these measures, we benchmark two popular families of post-hoc explanation models: local methods, which imitate the boundary of a blackbox around one instance~\cite{ribeiro2016should,lundberg2017unified,ribeiro2018anchors}, and global methods, which imitate the blackbox across all instances \cite{rudin2014algorithms,lakkaraju2019faithful}. We also motivate fidelity gap measurements by showing mathematically that measuring fidelity gaps across subgroups directly connects with prior work on fairness preservation for explainability \cite{dai2021will}. 
With a comprehensive audit of explanation fairness, we find that significant fidelity gaps exist between subgroups.

A popular way to train fairer models is through robust optimization \cite{liu2021just,rahimian2019distributionally,sagawa2019distributionally}.
To see how robust training impacts large fidelity gaps, we also study a simple technique for retraining explainability methods to improve their fairness. 
We also study potential causes for these fidelity gaps, and highlight mechanisms by which group information can indirectly be used in post-hoc explanations as an important contributing factor. Lastly, we assess the impact of the observed fidelity gaps on real-world decision-making accuracy with a carefully designed simulation study. The major findings of our evaluation are as follows:
 

\textbf{Explanation fidelity varies significantly between subgroups:} We find that fidelity gaps grow up to 7\% between subgroups in our experiments using four popular datasets. In comparison to average fidelity across all data points, the fidelity of explanations for disadvantaged groups is often significantly lower (up to 21\%).
These findings indicate that judging the quality of explanations by their \textit{average} fidelity alone---a common approach---overestimates explanation quality for some subgroups, potentially leading to worse downstream decision making.
This effect is illustrated in Fig.~\ref{fig:local_global_expls}.

\textbf{Balanced and robust training can reduce but not eliminate fidelity gaps:} We use robust training by adaptively reweighting or balancing groups in training data while training explanation models. This turns out to be a promising direction: fidelity gaps improve across subgroups, though this depends on both the dataset and exact method utilized.

\textbf{Fidelity gaps have an impact on decision-making in the real-world:} Using a simulation study, we find that larger fidelity gaps may lead to disparities in decision making accuracy for different subgroups. This implies that ignoring fidelity gaps between subgroups can have detrimental effects to members of protected groups.

Finally, we categorize and discuss promising directions for evaluating and improving post-hoc explainability methods. In summary, our work is a step towards training fair and reliable explanation models.

\section{Related Work}

\subsection{Explainable Machine Learning}

While ML models achieve outstanding performance, users often find them too complex to trust in practice~\cite{dietvorst2015algorithm}.
To make such \textit{blackbox} models more useful, users require that they be understandable, often 
due to laws \cite{bibal2021legal} or preference \cite{holzinger2018machine, roscher2020explainable, bhatt2020explainable}.
To fill this gap, recent approaches ``explain'' a blackbox model's behavior after it is trained \cite{dovsilovic2018explainable, burkart2021survey}. These \textit{post-hoc} explainability methods are now used in safety-critical applications like healthcare \cite{ahmad2018interpretable} and finance \cite{bussmann2021explainable}.

Several explainability methods are increasingly-popular because they make no assumptions about a blackbox model's architecture \cite{ribeiro2016model}, also known as model-agnostic. 
In contrast, some methods are designed exclusively for deep learning, requiring their internal structure and gradients \cite{shrikumar2017learning, shrikumar2017learning, bach2015pixel,li2015visualizing, karpathy2015visualizing}.
We consider model-agnostic methods, which can be used for a wider family of blackbox models, including deep learning.

Model-agnostic explainability methods are primarily either \textit{local} or \textit{global} \cite{du2019techniques}.
\textit{Local} methods justify one model prediction at a time, typically by approximating the decision boundary around one data point \cite{plumb2018model, botari2020melime, lundberg2017unified, ribeiro2018anchors, rathi2019generating, ribeiro2016should}. Then, the weights learned by the local models are used to rationalize the blackbox model's prediction. Some of the best-known local methods are LIME \cite{ribeiro2016should}, which learns a sparse linear classifier on a dataset of perturbed samples, and SHAP \cite{lundberg2017unified}, which uses feature-wise Shapley values \cite{roth1988shapley}. 
\textit{Global} methods, on the other hand, train interpretable surrogate models of the blackbox model's behavior on an entire dataset, which is then used in lieu of the blackbox. 
These methods primarily use tree-based models \cite{lundberg2020local}, rule lists \cite{lakkaraju2019faithful, puri2017magix}, sparse linear models \cite{ustun2016supersparse, zhang2021learning}, and generalized additive models \cite{lou2012intelligible} as surrogates. 

\subsubsection{Explainable ML in Safety-critical Settings} 
The need for explainability in safety-critical applications is a nebulous and contested topic for several reasons:

\noindent\textbf{Interpretability vs Explainability.} Some prior works advocate for interpretability over explainability~\cite{rudin2019stop,mittelstadt2019explaining,ghassemi2021false}. An explanation model without perfect fidelity is by definition incorrect for some data points~\cite{rudin2019stop}. Our work extends this point; these errors can occur for some groups more than others. Since explanations influence trust~\cite{buccinca2020proxy}, it is important to conduct user studies on the impacts on aversion algorithmic advice~\cite{logg2019algorithm} and over-reliance on algorithmic advice~\cite{dijkstra1998persuasiveness}.

\noindent\textbf{Anchoring Effects.} Explanations can fool people into trusting incorrect models~\cite{bansal2021most,poursabzi2021manipulating}. For example, \citet{poursabzi2021manipulating} find that when people are shown explanations from a bad model, they become more likely to trust the model, even when it is clearly wrong.
In cases like this, people use the explanations while judging the quality of the blackbox models, even though the explanations themselves can be misleading \cite{bansal2021does}.

\noindent\textbf{Mismatched end-user and model-designer goals.}
Many explainability methods aim to assist model debugging, while non-engineer users only choose when to accept a blackbox's decisions \cite{ribeiro2016model,krishnan2017palm,pradhan2021interpretable}.
This mismatch can have downfalls.
For instance, \citet{buccinca2021trust} find that the explanations people find most useful are also the ones they trust \textit{incorrectly}.
Resolving this mismatch requires goal-aware explainability methods along with education to ensure end-users are properly trained in using these methods.

\subsubsection{Desiderata for Post-hoc Explanations}
Most post-hoc explainability methods have three goals:

\indent\textbf{Reliability.} Explanations must be accurate for the right reasons. People often trust explained models \cite{poursabzi2021manipulating,ross2017right}, so ensuring that explanations are faithful to the original model and not simply easy-to-rationalize is essential \cite{ghassemi2021false}. 
    
\indent\textbf{Robustness.} Explanation models should not overfit to spurious patterns in the data~\cite{ghorbani2019interpretation,lakkaraju2020fool} and must be robust in the presence of small distribution shifts at test time~\cite{lakkaraju2020fool}.
    
\indent\textbf{Simplicity.} Models should be sparse, and leave little room for effects for human cognitive biases such as the anchoring effect~\cite{poursabzi2021manipulating}. Ideal explanations will highlight only the key information needed to understand a model's behavior, encouraging users to engage with explanations in predictable ways~\cite{buccinca2021trust}. However, there is often a trade-off between an explanation's faithfulness and its simplicity~\cite{lakkaraju2017interpretable}. Recent work on cognitive forcing---where users explicitly interact and understand explanations---appears to be a promising direction to address this trade-off~\cite{buccinca2021trust}.

Along with other recent efforts~\cite{ghorbani2019interpretation,lakkaraju2020fool,poursabzi2021manipulating,bansal2021does}, we promote a fourth goal: \textbf{Fairness}.
Explanation quality should not depend on group membership.
We find that this requirement is not yet satisfied by popular explainability methods.

\subsection{Algorithmic Fairness}
\label{sec:algorthmic_fairness_background}
Formalizing fairness is a flourishing research area
~\cite{mehrabi2021survey, chen2018classifier, chouldechova2017fair, berk2017impact, locatello2019challenging, chouldechova2018frontiers, hardt2016equalityofopportunity, zemel2013learning, zafar2017fairness,dwork2011fairnessthroughawareness}.
Recent works define fairness at either the \textit{individual}- or \textit{group}-level.
Individual fairness~\cite{dwork2011fairnessthroughawareness} requires similar predictions for similar individuals; group fairness requires similar predictions for different groups (sex or race, for example).
We consider group-level fairness for binary classification, which we quantify using demographic parity (DP) gap \cite{hardt2016equalityofopportunity,pleiss2017}, a standard group-fairness metric.
We describe this metric probabilistically, allowing calculation of gaps across groups: $\text{DP} = E[\hat{Y}\vert A = a] - E[\hat{Y}\vert A = b] \quad \forall a,b \in A$, where $\hat{Y}$ is a predictor and its DP is measured with respect to attribute $A$.

There are three main strategies for encouraging group fairness 
~\citep{caton2020fairness}: pre-processing data to find less-biased representations~\cite{paul2021generalizing};
enforcing fairness while training a model, typically through regularization~\cite{madras2018learning,zhang2018mitigating};
and altering a model's predictions to satisfy fairness constraints after it is trained~\citep{hardt2016equalityofopportunity, agarwal2018reductions, pleiss2017, chen2020}.
In this paper, we utilize the inprocessing  method proposed by ~\citet{zhang2018mitigating} for training fair blackbox models. Further, recent work has demonstrated that group-robust training can increase fairness by improving the worst-group accuracy~\cite{sagawa2020groupdro}. 

\subsection{Bias in Model Compression and Risks of Fairwashing}
Several recent works study the effects of model and data compression on fairness~\cite{samadi2018price,hooker2020characterising}. For example, ~\citet{samadi2018price} observe that reconstruction error associated with data dimensionality reduction via principal component analysis is higher for some populations.
\citet{hooker2020characterising} show that average accuracy after ML model compression 
hides disproportionately high errors on a small subset of examples.
In a similar vein, we study post-hoc explanation models, which are often compressed blackbox models, and assess how they transmit bias. Another related topic is ``fairwashing'': the act of overlooking a model's unfair behavior by rationalizing its predictions via explanations \cite{aivodji2019fairwashing}. Our paper instead considers fairness in how well explanation models imitate blackbox models (rather than the ground-truth), regardless of blackbox model fairness. In concurrent work, ~\citet{dai2022} showed that explanation quality may differ between subgroups, further validating our findings. However, the metrics and methodological focus in their work is on feature-based variations to fidelity. We encourage readers to review their work for an alternate approach to measuring the fairness of explanations.


\section{Measuring the Fairness of Explanations}
\label{sec:background}
Here, we introduce metrics for measuring \emph{fairness of explanation models} or fidelity gaps across subgroups. 

\subsection{Notation}
Consider a dataset $\mathcal{D} = \{(\bm{x}_i, y_i)\}_{i=1}^n$ that contains $n$ training data points. $\bm{x}_i \in \mathbb{R}^d$ is the $d$-dimensional feature vector of the $i$-th data point in $\mathcal{D}$ and $y_i \in \{0, 1\}$ is its associated binary label.
We assume binary classification for simplicity.
Let $g_i \in \{1, \dots, G\}$ be a variable defining group membership with respect to the protected attribute for every data point 
for a total of $G$ subgroups. In all cases, $g$ serves as auxiliary information and is not used during any model training, unless specified otherwise.
A blackbox classifier $B: \mathbb{R}^d \rightarrow \{0, 1\}$ predicts one binary label per input $x$.
Given classifier $B$, we wish to explain its prediction given some query point $\bm{x}^*$.
To achieve this, an explanation model $E$ is chosen from a set of interpretable models (\textit{e.g.}, linear models or decision trees).
Then, $E$ is trained to \textit{imitate} $B$ either locally (for the feature space near $\bx^*$) or globally (for all data points in $\mathcal{D})$.




\subsection{Fidelity of Explanations}

Given a blackbox $B$ and explainability model $E$, we seek to describe how well $E$ approximates $B$'s behavior. Fidelity, as detailed in Definition \ref{bg:def:fidelity} below, is a powerful measure for this approximation error \cite{aivodji2021characterizing,lakkaraju2017interpretable,lakkaraju2019faithful}, though it disregards group information. 

\begin{definition}[Explanation Fidelity~\cite{craven1995extracting}]
\label{bg:def:fidelity}
Given blackbox model $B$ and explanation model $E$, the \textit{explanation fidelity} on data points $(x_i,y_i)_{i=1}^N$ is
$\frac{1}{N}\sum_{i=1}^{N} L(B(x_i), E(x_i))$,
where $L$ is a performance metric.
\end{definition}
For $L$, we use accuracy, AUROC\footnote{AUROC cannot be written directly as a sum but we slightly abuse notation for readability.}, and mean error,
denoted as $Fidelity^{Acc}$, $Fidelity^{AUROC}$, $Fidelity^{Err}$, respectively.

In the following sections, we build up to a definition of explanation fidelity that considers group information.
First, we motivate the need for a metric that measures fidelity across groups (Section~\ref{sec:motivation}), then define two new notions for measuring the fairness of explanations (Section~\ref{sec:metrics}).

\subsection{Fidelity Gaps are Critical to Fairness Preservation}\label{sec:motivation}

\citet{dai2021will} recently introduced \textit{fairness preservation} in surrogate explanation models. 
Fairness is \textit{preserved} when the fairness properties of the blackbox model and explanation model are identical.
For example, consider Figure~\ref{fig:local_global_expls}.
A linear explanation $E$ is a high-fidelity approximation of the blackbox $B$'s decisions for one group ($\triangle$), but not the other ($\square$). Here, $B$ seems unfair in predicting the ``unhealthy" class 
for the two groups. Meanwhile, $E$ appears fairer.
In this example, $B$'s degree of (un)fairness---the demographic parity gap---is not \emph{preserved} by the explanation model.
For demographic parity, fairness preservation in explanations implies that $B$ and $E$ should have similar DP Gaps (Section~\ref{sec:algorthmic_fairness_background}).  

To reliably judge a blackbox's fairness using only its post-hoc explanations, preserving fairness is essential.
If fairness is preserved, then when an explanation seems unfair, we can be confident that the blackbox model is likely similarly unfair as well.
Next, we prove that fairness preservation is directly linked to fidelity gaps across subgroups. While ~\citet{dai2021will} briefly intuit that fairness preservation impacts  explanation fidelity via an illustratory example, only a group-conditional blackbox model's decision boundary under imbalanced group sizes is considered. In contrast, we do not make any assumptions about the relative sizes of groups or group-dependence of blackbox model and instead show that fairness preservation is related to fidelity gaps (more so than overall explanation fidelity).


\subsubsection{Fidelity Gaps are related to Fairness Preservation}

\begin{theorem}\label{thm:thm}
Let \(E\) be a post-hoc explanation model trained to imitate predictions of blackbox model \(B\), and mean residual error for a set of $N$ data points in dataset $\mathcal{D}$ is $\frac{1}{N}\sum_{x \in \mathcal{D}}(E(x)-B(x))$. Then, the difference between the Demographic Parity Gaps of \(E\) and \(B\), both with respect to binary valued-protected attribute $g$, is equal to the difference in mean residual error of data points with $g=1$ and $g=0$.
\end{theorem}

The full proof of this theorem is in Appendix~\ref{sec:proof}; the key idea is to expand $E(x_i)=B(x_i)+\epsilon_i$ where $\epsilon_i$ is a residual for each point $x_i$. This is valid when $E$ is trained to imitate $B$ with high fidelity (\textit{e.g.}, minimizing mean squared error or cross-entropy loss). We also empirically validate this theorem on explanation models in Appendix~\ref{sec:empirical_validation}.

From Theorem \ref{thm:thm}, a \emph{sufficient} condition for DP, as computed over instances $x_{i}$ and their local linear classifiers $E_{i}(x_{i})$ or a global model $E$ (where $E_{i}=E \forall i$) is ensuring that the mean residual errors for each group is comparable. This is the same as low fidelity gaps across subgroups where $L$ is the mean error. Note that this does not correspond to mean absolute difference between predictions of $E$ and $B$, but instead their mean difference. Theorems of similar form could be derived for other group fairness definitions, but the $\epsilon$ values and data points considered would depend on the ground truth as well (e.g., for equal-opportunity, the $\epsilon$ difference would only contain terms for data points with positive-class ground truth).  
With this motivation in mind, we next introduce two new metrics that measure fidelity gaps across subgroups.



\subsection{Measuring the Fairness of Explainability Methods}\label{sec:metrics}

Building on the definition of average fidelity across groups (Defn.~\ref{bg:def:fidelity}), we introduce two new measurements for the fairness of explanation models by evaluating their fidelity gaps between subgroups. The first metric (Definition~\ref{bg:def:subgroup_avg_gap}) addresses the question: by what degree would relying on the average fidelity alone be detrimental to subgroups of data? The second metric estimates the mean difference in fidelity of explanations between subgroups of data (Definition~\ref{bg:def:subgroup_max_gap}).


Inspired by past work~\cite{liu2021just,craven1995extracting}, the \textit{maximum fidelity gap from average} (Definition \ref{bg:def:subgroup_avg_gap}) computes the difference between the overall, average fidelity and the worst-case subgroup fidelity.
This way, we quantify the maximum degree to which an explanation model's fidelity is lower for disadvantaged groups compared to the average across all subgroups.

\begin{definition}[Maximum Fidelity Gap from Average: $\Delta_{L}$]
\label{bg:def:subgroup_avg_gap}
Let the maximum fidelity gap from average be
\[\Delta_{L}=\max_{j}\left[\frac{1}{N} \sum_{i=1}^{N} L(B(x_i), E(x_i)) - \frac{1}{N_j} \sum_{i : g_i^j=1}L(B(x_i), E(x_i))\right],\]
where $g_i^j=1$ denotes that point $x_i$ belongs to the $j$th subgroup defined by a specific protected attribute $g$ (\textit{e.g.} data points from females), and $N_j$ is the number of data points with $g^j=1$.
\end{definition}

Next, the mean fidelity gap amongst subgroups (Definition \ref{bg:def:subgroup_max_gap}) computes how much an explanation model's fidelity differs over subgroups.
Here, we only consider groups defined by the same sensitive attribute (\textit{e.g.} $g^k$ is male, $g^j$ is female).

\begin{definition}[Mean Fidelity Gap Amongst Subgroups: $\Delta^{group}_{L}$]
\label{bg:def:subgroup_max_gap}
Let the mean fidelity gap amongst subgroups be
\[\Delta_{L}^{group}=\frac{2}{G(G-1)}\sum_{k=1}^{G}\sum_{j=k+1}^{G}\left|\frac{1}{N_k} \sum_{i : g_i^k=1}L(B(x_i), E(x_i)) - \frac{1}{N_j} \sum_{i : g_i^j=1}L(B(x_i), E(x_i))\right|,\]
where $g^j$ denotes the $j^{th}$ subgroup defined by a specific sensitive attribute (\textit{e.g.} datapoints from females), and $N_j$ is the number of datapoints in $g^j$.
\end{definition}

Similar to average fidelity, we choose $L$ to be Accuracy, AUROC, and Mean Error for both fidelity gap measurements (e.g., $\Delta_{AUROC}$ and $\Delta^{group}_{AUROC}$). In all cases, we do not consider intersectional groups due to sample size concerns.  


\subsection{Experiments Overview}
Since fidelity gaps across subgroups are closely linked to fairness preservation and risks of fairwashing, we design experiments to audit this quantity. We conduct the following experiments in the sections below\footnote{Code: \url{https://github.com/MLforHealth/ExplanationsSubpopulations}}:

\textbf{Measuring Fidelity Gaps Between Subgroups:} We measure fidelity gaps using metrics defined in ~\ref{bg:def:subgroup_avg_gap} and ~\ref{bg:def:subgroup_max_gap} for four different post-hoc explanation models, and two different blackbox model classes. The aim of this experiment is to study the presence and degree of fidelity gaps in standard explainability methods (Section~\ref{sec:fidelity_gaps_exps}).

\textbf{Assessing the Impact of Robust Training:} We use robust training strategies to train explanation models, and repeat the fidelity gap audits to study if robust training can provide reduced fidelity gaps (Section~\ref{sec:robust_training}).

\textbf{Studying Possible Causes for Fidelity Gaps}: We analyze the impact of blackbox fairness and presence of protected attribute information in feature representations on the fidelity gap  (Section~\ref{sec:fair_fidelity_gaps}).

\textbf{Simulation Showing Impact of Fidelity Gaps}: We conduct a simulation and study the quality of decisions made for groups to examine the impacts of unfair explanation models on real-world decision making (Section~\ref{sec:real_world}).

\begin{table}[]
\begin{tabular}{llccccc}
\toprule
\textbf{Dataset} & \textbf{Outcome Variable} & \textbf{n} & $\bm{d}$ & $\bm{d}'$ & \textbf{Protected Attribute ($g$)}   \\ \midrule
\texttt{adult} \cite{Dua:2019} & Income $>$ 50K            & 48,842      & 9  & 33  & Sex (2 groups) \\
\texttt{lsac} \cite{wightman1998lsac} & Student passes the bar    & 20,427      & 8          & 14 & Race (5 groups) \\
\texttt{mimic} \cite{harutyunyan2019multitask} & Patient dies in ICU       & 21,139      & 49         & 49 & Sex (2 groups)\\
\texttt{recidivism} \cite{propublica2019} & Defendant re-offends       & 6,150       & 6         &  7 & Race (2 groups)  \\ \bottomrule
\end{tabular}
\caption{Binary classification datasets used in our experiments. $n$ is the number of samples, $d$ is the number of variables in the original dataset, and $d'$ is the number of features after one-hot encoding categorical variables. \label{tab:datasets}}
\end{table}


\section{Explanation Fidelity Varies Significantly between Subgroups}
\label{sec:fidelity_gaps_exps}
Experimentally, we find that fidelity gaps indeed vary by group in many settings. To show this, we train four post-hoc explainability methods (two local, two global) to explain two different blackbox models trained on the four standard fairness benchmark tabular datasets described in Table 1. Following ~\citet{aivodji2021characterizing}, we randomly split each dataset into four subsets: a training set for blackbox models (50\%), a training set for explanation models (30\%), a validation set for explanation models (10\%), and a held-out test set for evaluating both blackbox and explanation models (10\%). For each dataset, we train both a Neural Network (NN) and a Logistic Regression (LR) model to serve as blackboxes. See Section~\ref{sec:training_regimes} in the Appendix for details on the training regimes, hyperparameter settings, and evaluation metrics for each. In the following sections, we describe the explainability models and fidelity gaps observed.

\subsection{Local Explanation Models}
Local explanation models explain individual predictions from classifiers by learning an interpretable model locally around each prediction. In our experiments, we consider LIME \cite{ribeiro2016should,ribeiro2016model} and SHAP \cite{lundberg2017unified}, which are popular methods that use linear models to elicit each feature's contribution to the blackbox model's prediction. More details are in Appendix \ref{sec:explanation_model_descriptions}.

\textbf{Experiment Setup.} We measure fidelity gaps between subgroups using the two key metrics introduced in Section \ref{sec:metrics} (see Definitions ~\ref{bg:def:subgroup_avg_gap} and ~\ref{bg:def:subgroup_max_gap}).
For each, we select three performance measures: Accuracy ($\Delta_\text{Acc.}^\text{group}$) following prior work \cite{aivodji2019fairwashing}, mean residual error ($\Delta_\text{Err.}^\text{group}$), and also include AUROC ($\Delta_\text{AUROC}^\text{group}$) as a threshold-independent metric. A full table with all metrics can also be found in the Appendix (Section~\ref{sec:full_gaps_tables}). 
For accuracy, we use a threshold of 0.5.
Since the four datasets are imbalanced, we use AUROC for model selection while tuning all hyperparameters.
Non-zero fidelity gaps indicate disparities across groups in the explanation models.  


\textbf{Results.}
First, we find that LIME disproportionately favors different groups, as shown in Table \ref{tab:local}, where the maximum accuracy gap ranges from 0.1-21.4\%.
This confirms that explanation quality can dramatically differ by subgroup, even without access to group-membership data.
Furthermore, the AUROC/Accuracy between protected groups also ranges significantly (0-6.6\%/0.3-20.6\%), indicating that some members of protected groups are disadvantaged in terms of explanations. Hence, when explanations are judged to be ``high quality" based on average fidelity, it might be misleading and lead to errors in decision-making. Bolded non-zero fidelity gaps are also significantly greater than $0$ with a one-sided Wilcoxon signed-rank test at $p<0.05$. 

\begin{figure}[ht!]
    \centering
    \begin{subfigure}{.45\textwidth}
        \includegraphics[width=1\linewidth]{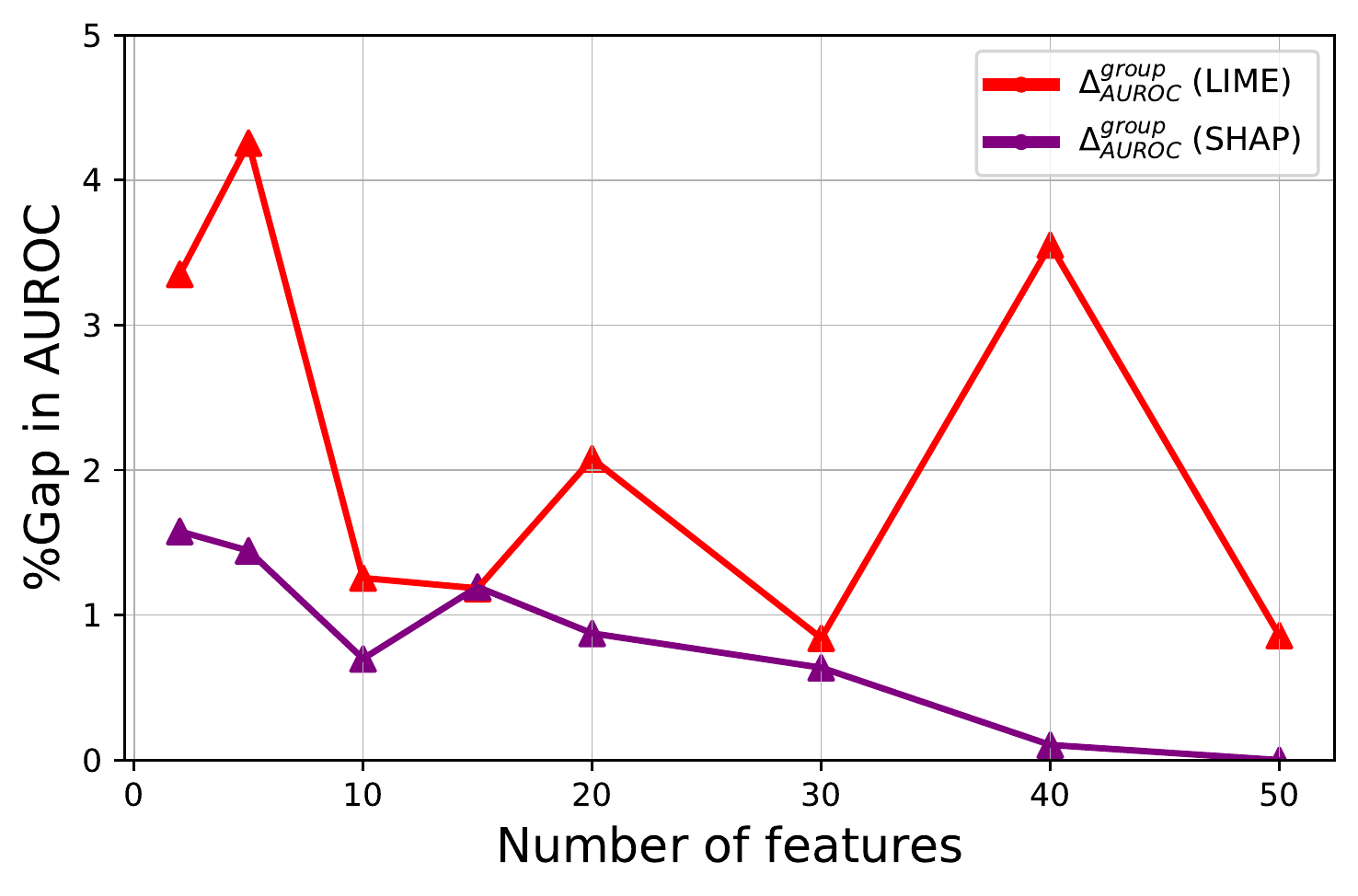}
        \caption{Local Explanations}
    \end{subfigure} 
    \begin{subfigure}{.45\textwidth}
        \includegraphics[width=1\linewidth]{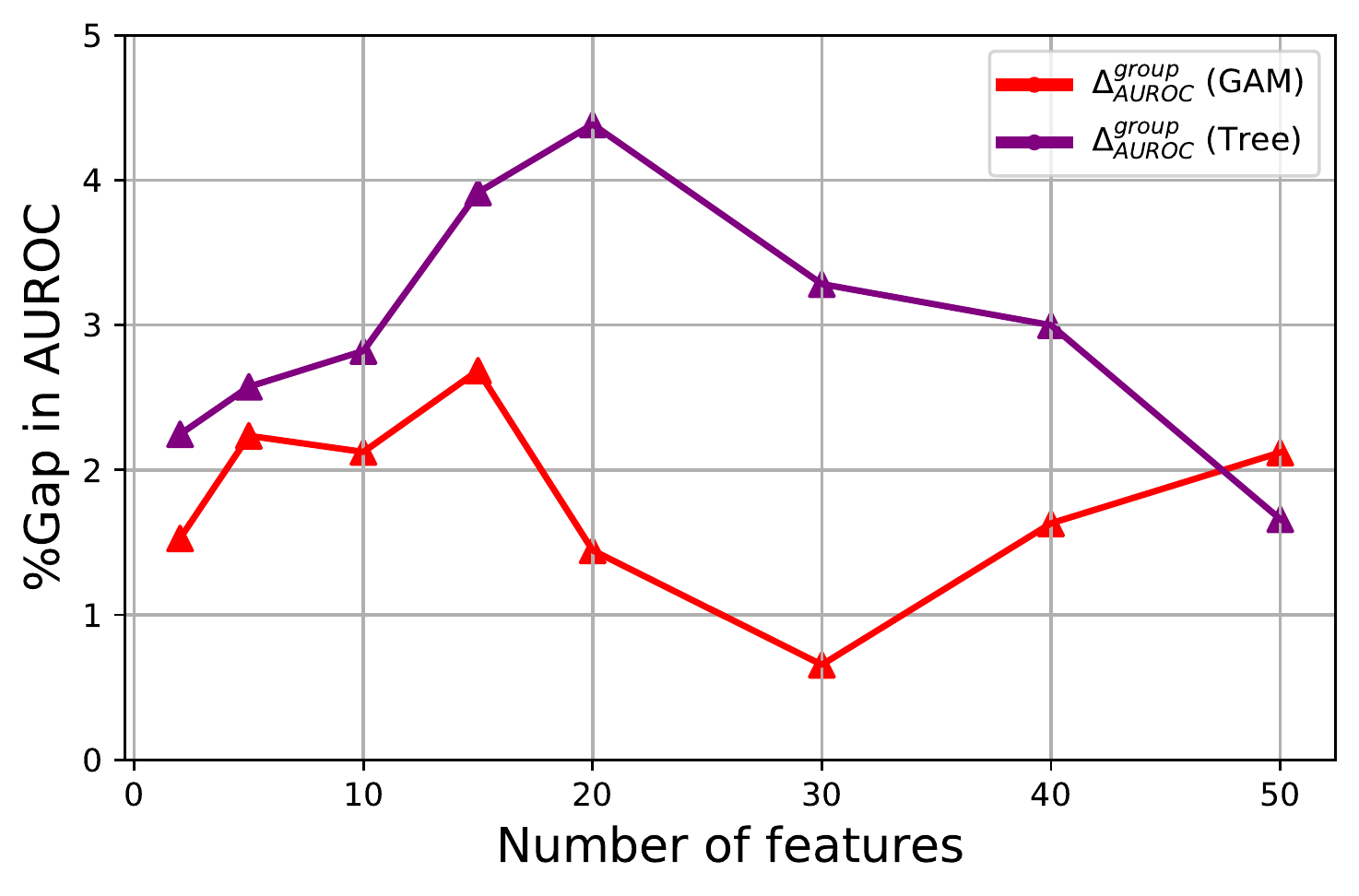} 
        \caption{Global Explanations}
    \end{subfigure}
    \caption{The effect of varying the number of features on fidelity gaps using the \texttt{mimic} dataset with a neural network blackbox model. For (a) local explanation models, using fewer features leads to worse fidelity gaps. We observe larger fidelity gaps across subgroups with sparser models, i.e., fewer features in (a) local explanation models. For (b) global explanation models, the gap varies with number of features. We also observe similar trends on other datasets (see Appendix~\ref{sec:varying_features_plots_all}).
    }\label{fig:auroc_gaps}
    \vspace{-5mm}
\end{figure}

Second, as expected, SHAP's gaps are consistently zero.
This is because the blackbox and explanation models are trained using identical features, in which case consistency is guaranteed~\cite{lundberg2017unified}.
However, using a subset of features to train the explanation model often leads to more useful explanations~\cite{wong2021leveraging}.
This increases the gaps significantly, as shown in Figure \ref{fig:auroc_gaps}, indicating that SHAP can also suffer from significant gaps in fidelity when used in practice. Since LIME considers sparsity as well, we also run this same experiment for LIME and find that fewer features are indeed associated with larger fidelity gaps (Fig.\ref{fig:auroc_gaps}). Increasing sparsity is a common approach in training explanation models and these experiments indicate that this technique alone may contribute to substantially worse fidelity gaps.

Third, we observe that the fidelity gaps in AUROC tend to be lower for the logistic regression blackbox, possibly because the linearity of the local surrogate models matches logistic regression better than the neural network. Note that the overall fidelity of all models are greater than 85\% (see Table~\ref{tab:full_gaps_local} in the Appendix).

\begin{table*}

\begin{tabular}{lccccc}
\toprule
Dataset & Blackbox Classifier  & $\Delta_\text{Acc.}$ & $\Delta^\text{group}_\text{AUROC}$ & $\Delta^\text{group}_\text{Acc.}$ & $\Delta^\text{group}_\text{Err.}$ \\
\midrule
\multirow{2}{*}{\texttt{adult}} &    Logistic Regression &     
\textbf{0.8\% ± 0.0\%} &    0.1\% ± 0.0\% &       \textbf{2.4\% ± 0.1\%} &      \textbf{1.9\% ± 0.0\%} \\
 &             Neural Network &              
\textbf{6.9\% ± 0.7\%} &    \textbf{3.0\% ± 1.2\%} &      \textbf{20.6\% ± 2.0\%} &      \textbf{0.8\% ± 0.5\%} \\
\midrule
\multirow{2}{*}{\texttt{lsac}} &             Logistic Regression &             
\textbf{2.0\% ± 1.0\%} &    0.0\% ± 0.0\% &       \textbf{1.5\% ± 0.5\%} &      \textbf{1.5\% ± 0.1\%} \\
 &             Neural Network &              
 \textbf{21.4\% ± 4.4\%} &    \textbf{6.6\% ± 1.2\%} &      \textbf{12.2\% ± 2.2\%} &     \textbf{3.8\% ± 1.2\%} \\
 \midrule
\multirow{2}{*}{\texttt{mimic}} &             Logistic Regression &             
\textbf{0.4\% ± 0.6\%} &    \textbf{3.0\% ± 1.8\%} &       \textbf{1.1\% ± 0.3\%} &      \textbf{2.0\% ± 0.1\%} \\

 &             Neural Network &              
\textbf{0.8\% ± 0.4\%} &   \textbf{ 1.7\% ± 1.5\%} &       \textbf{1.4\% ± 0.7\%} &      \textbf{1.7\% ± 0.5\%} \\
 \midrule
\multirow{2}{*}{\texttt{recidivism}} &             Logistic Regression &              
0.1\% ± 0.1\% &    0.0\% ± 0.0\% &       0.3\% ± 0.2\% &      0.3\% ± 0.0\% \\
 &             Neural Network &              
 \textbf{0.9\% ± 0.3\%} &    \textbf{0.7\% ± 0.3\%} &       \textbf{2.4\% ± 0.7\%} &      \textbf{1.1\% ± 0.1\%} \\
 \bottomrule
\end{tabular}
\caption{Performance fidelity gaps across subgroups for \textit{LIME local explanations} using all available features. ± denotes standard deviation computed over 5 replications. Fidelity gaps are significant (one-sided Wilcoxon signed-rank tests at $p<0.05$; marked in \textbf{bold}) between all five groups in the \texttt{lsac} dataset, and between two sensitive groups in other three datasets. $\Delta_{Acc.}$ denotes the maximum fidelity gap of subgroups from average (in terms of accuracy at $0.5$ threshold), and $\Delta^{group}_{m}$ is the mean fidelity gap between subgroups using metric $m$.}\label{tab:local}
\vspace{-10mm}
\end{table*}

\subsection{Global Explanation Models}

Global explanation methods train one new surrogate model that approximates the behavior of a blackbox model.
This surrogate model should itself be easily understood, and can then be used instead of the blackbox at test time (more background in Appendix \ref{sec:explanation_model_descriptions}).

\textbf{Experiment Setup.} In this experiment, we generate global explanations using two popular choices of interpretable surrogate models: Generalized Additive Model (GAM) \cite{hastie2017generalized} and a sparse decision tree (Tree)~\cite{pedregosa2011scikit}.
GAM combines linear models of different variables during explanation \cite{tan2018distill}, while Tree uses a low-depth, sparse, decision tree.
We evaluate the fidelity of each global method with the original blackbox and compare across subgroups.
As with the local methods, we use both Accuracy and AUROC to evaluate fidelity gaps.

\textbf{Results.}
First, we find that the fidelity gap between subgroups differs substantially from the average for the global explanation models, as shown in Table \ref{tab:global} where the accuracy gap ranges from 0-13.5\%. We again observe that AUROC and accuracy vary substantially between protected subgroups (0-8.1\% and 0.1-7.4\% for protected groups such as sex and race groups in each dataset).
This is especially true for more imbalanced subgroup proportions: having more subgroup categories leads to more disadvantage in protected groups, particularly when the classes are imbalanced themselves (e.g. \texttt{lsac}). 

Second, we find that using fewer features (e.g., 15 in Fig.~\ref{fig:auroc_gaps}) may lead to larger gaps in performance between subgroups in sparse decision trees (Trees), bolstering prior findings on training trustworthy models \cite{chang2021interpretable}.
Hence, the gaps shown in Tables \ref{tab:local} and \ref{tab:global} are likely underestimates when using fewer dimensions in explanation models, which is common.
Interestingly, the subgroup with the lowest-quality explanations is not always the minority subgroup---which may be the most disadvantaged---in the datasets for fair ML. We expand this finding in Table \ref{tab:worst_case_groups} in the Appendix. 
Additionally, we see that subgroup gaps occur even after training blackbox models with a balanced number of data points from each subgroup for both global and local explanation models (see Table~\ref{tab:balanced_global}). 



\begin{table*}[t]
\begin{tabular}{lccccccc}
\toprule
Dataset & Blackbox Classifier & Expl. Model & $\Delta_\text{Acc.}$ & $\Delta^\text{group}_\text{AUROC}$ & $\Delta^\text{group}_\text{Acc.}$ & $\Delta^\text{group}_\text{Err.}$ \\
\midrule
\multirow{4}{*}{\texttt{adult}} &             Logistic Regression &               GAM & 0.1\% ± 0.0\% &    0.0\% ± 0.0\% &       0.3\% ± 0.0\% &      0.1\% ± 0.0\% \\
 &             Logistic Regression &              Tree &  \textbf{1.5\% ± 0.1\%} &    \textbf{2.9\% ± 0.4\%} &       \textbf{4.5\% ± 0.2\%} &      \textbf{1.1\% ± 0.1\%} \\
 &             Neural Network &               GAM &  
 \textbf{0.8\% ± 0.2\%} &    \textbf{0.5\% ± 0.3\%} &       \textbf{2.4\% ± 0.5\%} &      0.3\% ± 0.2\% \\
 &             Neural Network &              Tree &  \textbf{1.1\% ± 0.1\%} &    \textbf{0.6\% ± 0.4\%} &       \textbf{3.4\% ± 0.2\%} &      \textbf{0.5\% ± 0.4\%} \\
 \midrule
\multirow{4}{*}{\texttt{lsac}} &             Logistic Regression &               GAM &  
\textbf{0.9\% ± 0.9\%} &    0.0\% ± 0.0\% &       \textbf{0.6\% ± 0.4\%} &      \textbf{0.7\% ± 0.3\%} \\
 &             Logistic Regression &              Tree &  
 \textbf{3.7\% ± 3.1\%} &    \textbf{1.1\% ± 0.4\%} &       \textbf{2.8\% ± 0.7\%} &      \textbf{1.8\% ± 0.5\%} \\
 &             Neural Network &               GAM &  
 \textbf{13.5\% ± 0.9\%} &    \textbf{5.2\% ± 1.2\%} &       \textbf{7.3\% ± 1.0\%} &      \textbf{3.9\% ± 2.6\%} \\
 &             Neural Network &              Tree &  
 \textbf{11.5\% ± 2.7\%} &    \textbf{5.8\% ± 2.1\%} &      \textbf{ 7.4\% ± 1.2\%} &      \textbf{4.9\% ± 2.0\%} \\
 \midrule
\multirow{4}{*}{\texttt{mimic}} &             Logistic Regression &               GAM &  
0.5\% ± 0.1\% &    0.4\% ± 0.1\% &       \textbf{0.9\% ± 0.1\%} &      0.4\% ± 0.2\% \\
 &             Logistic Regression &              Tree &  
 \textbf{0.6\% ± 0.0\%} &    \textbf{8.1\% ± 0.8\%} &       \textbf{1.2\% ± 0.1\%} &      \textbf{1.9\% ± 0.0\%} \\

 &             Neural Network &               GAM &  
 \textbf{1.2\% ± 0.3\%} &    \textbf{1.8\% ± 1.2\%} &       \textbf{2.2\% ± 0.6\%} &      \textbf{0.9\% ± 0.3\%} \\
 &             Neural Network &              Tree &  
 \textbf{1.1\% ± 0.5\%} &    \textbf{3.0\% ± 1.5\%} &       \textbf{2.0\% ± 0.9\%} &      \textbf{1.9\% ± 0.9\%} \\
 \midrule
\multirow{4}{*}{\texttt{recidivism}} &             Logistic Regression &               GAM &  0.1\% ± 0.0\% &    0.1\% ± 0.0\% &       0.3\% ± 0.0\% &      0.5\% ± 0.0\% \\
 &             Logistic Regression &              Tree &  0.0\% ± 0.0\% &    0.4\% ± 0.0\%&       0.1\% ± 0.0\% &     \textbf{ 1.2\% ± 0.0\%} \\
 &             Neural Network &               GAM &  
0.2\% ± 0.2\% &    0.4\% ± 0.2\% &       0.6\% ± 0.6\% &      \textbf{1.1\% ± 0.4\%} \\
 &             Neural Network &              Tree &  
 \textbf{0.9\% ± 0.3\%} &    \textbf{1.0\% ± 0.9\%} &       \textbf{2.3\% ± 0.7\%} &      \textbf{1.4\% ± 0.3\%} \\
 \bottomrule
\end{tabular}
\caption{Fidelity gaps across subgroups for \textit{global} explanation models GAM and Tree. ± denotes standard deviation computed over 5 replications. Fidelity gaps are significant (one-sided Wilcoxon signed-rank tests at $p<0.05$; marked in \textbf{bold}) for all five groups in the \texttt{lsac} dataset, and between two sensitive groups in the other three datasets with both global explanation models. $\Delta_{Acc.}$ denotes the maximum fidelity gap of subgroups from average (in terms of accuracy at $0.5$ threshold), and $\Delta^{group}_{m}$ is the mean fidelity gap between subgroups using metric $m$.}\label{tab:global}
\vspace{-10mm}
\end{table*} 

\section{Balanced and Robust Training Reduces Fidelity Gaps}\label{sec:robust_training}
Balanced and robust training methods could provide a path towards improving fidelity gaps, thereby learning fairer explanations~\cite{han2021balancing,adragna2020fairness}.
We showcase two such robust training methods, one for local methods and one for global methods.
For both cases, we choose hyperparameters that maximize the worst-case fidelity across all groups.\footnote{The overall fidelity is not significantly affected by either training approach.}
Ultimately, our experiments indicate that while robust training improves fidelity gaps sometimes, they remain largely pervasive. 


\subsection{Robust Local Explanation Models}
\textbf{Experiment Setup.} We train a more-robust version of LIME, using Just Train Twice (JTT)~\cite{liu2021just}, a two-stage training paradigm for training robust ML Models. First, we train an identification model via empirical risk minimization. Then, we extract its set of misclassified training examples. A final model is then trained by upsampling these misclassified examples, scaled by a hyperparameter $\lambda$. This reweighted loss is designed to make the second model more robust.
We use JTT to train LIME's local linear approximations, using linear models for both the identification and final models.

\textbf{Results.}
JTT successfully reduces gaps on three datasets with a NN blackbox model, as shown in Figure~\ref{fig:mod_gaps_lime}. 
Interestingly, this is not the case for the \texttt{recidivism} dataset, where JTT does not reduce the gaps and performs the same as standard training. With LR blackboxes (Figure~\ref{fig:mod_gaps_lime_all} in Appendix), the fidelity gaps are already small, so JTT is less impactful. However, non-zero gaps between 1-2\% still persist (e.g., NN blackbox on the \texttt{adult} dataset), indicating that the error-prone regions did not generalize to the test setting. Measuring fidelity gaps is therefore still critical, even if an explanation model is trained to be robust. 

\subsection{Robust Global Explanation Models}
\label{sec:robust_exps}

\textbf{Experiment Setup.} We next study balanced training for the global explanation method Tree.
We rebalance the explainability training sets for each dataset by randomly oversampling minority groups, a common approach for improving test error on minority subpopulations~\cite{wei2013role,han2021balancing}. This way, the training set in each case consists of an equal number of examples from each protected group.
Then, we train a Tree model to explain each blackbox model using these balanced datasets.

\begin{figure*}[t!]
\centering
\begin{subfigure}{0.48\textwidth}
  \includegraphics[width=\textwidth]{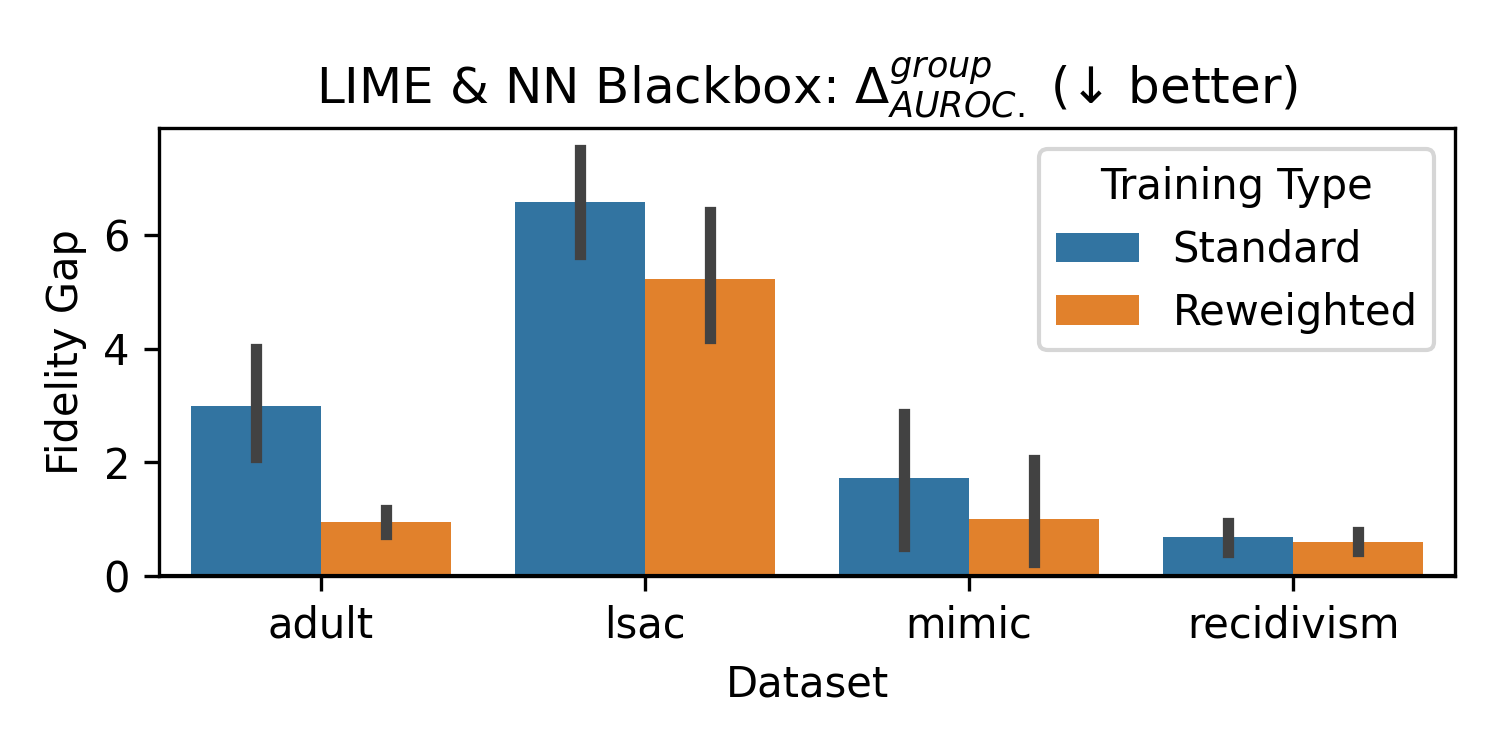}
  \caption{Local Explanations (LIME)}
\end{subfigure} 
\begin{subfigure}{0.48\textwidth}
  \includegraphics[width=\textwidth]{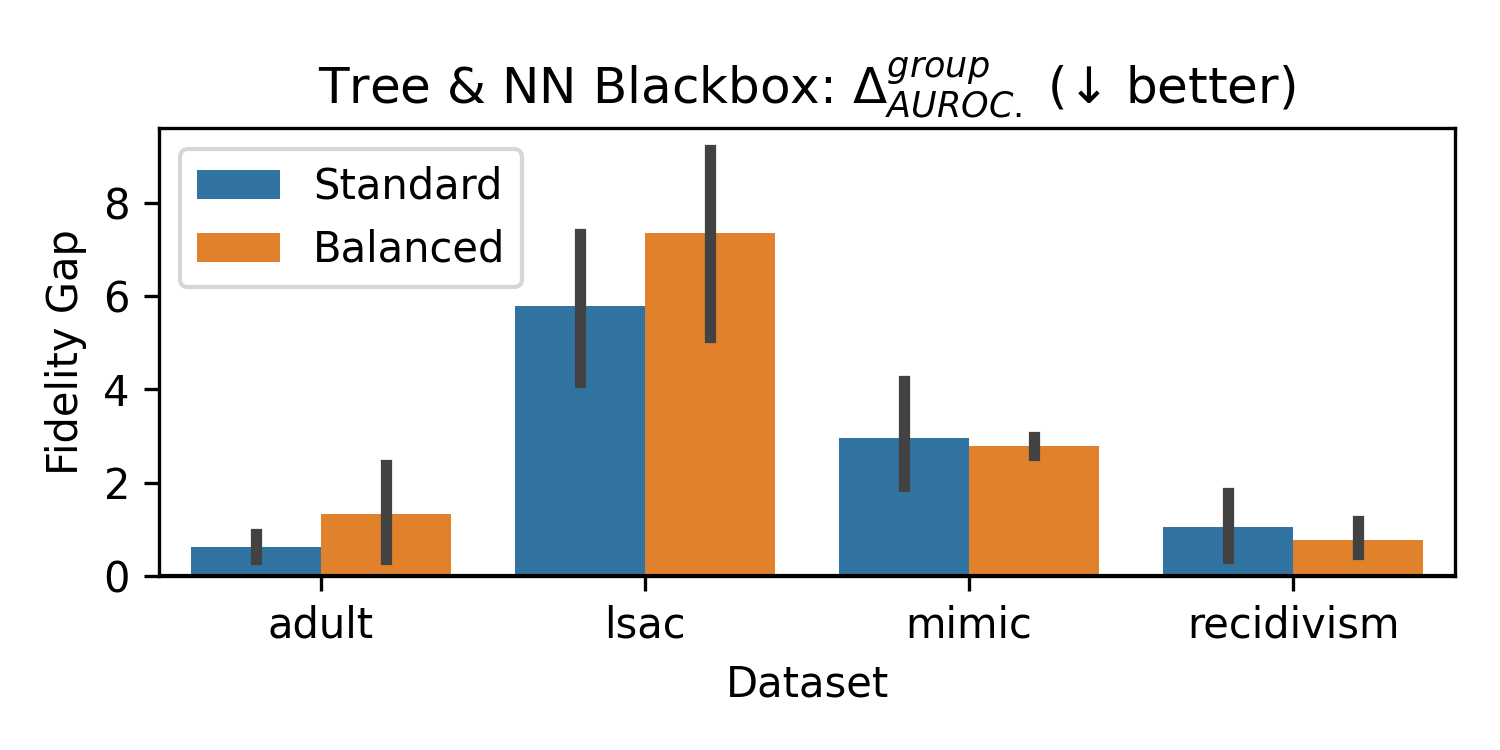}
  \caption{Global Explanations (Tree)}
\end{subfigure}
  \caption{AUROC Fidelity gaps across subgroups with and without robust training for (a) LIME and (b) Tree-based Models. Improvements are significant with robust training for \texttt{adult} dataset in (a) with a Wilcoxon-signed rank test at $p<0.1$ level. However, balanced training does not help for most datasets in (b). Error bars indicate 95\% confidence intervals.}\label{fig:mod_gaps_lime}
\end{figure*}

\textbf{Results.}
As shown in Figure \ref{fig:mod_gaps_lime}, we find that this common rebalancing approach to does not reduce gaps significantly across the board.
Still, some cases look more promising than others.
For example, \texttt{mimic} with NN which indicates this may be a fruitful direction for learning fairer explanations.
This is especially true for cases like \texttt{mimic} with LR, where rebalancing the training set increases the fidelity gap substantially (see Figure~\ref{fig:mod_gaps_lime_all} in Appendix). 

\section{On Possible Causes for Fidelity Gaps}
\label{sec:fair_fidelity_gaps}
 In our fidelity gap audits in prior sections, we noticed that the fidelity gaps are largest for the least-fair blackboxes  (\texttt{adult} and \texttt{lsac} datasets; Tables \ref{tab:global} and \ref{tab:local}). This indicates a potential relationship between the fairness of the blackbox and explainability models. To explore this further, we study the associations between blackbox fairness and fidelity gaps across subgroups in this section. First, we train fair models, and observe that significant non-zero fidelity gaps still persist. Second, we study if protected group information -- e.g. if a data point belongs to a Male or Female individual -- can be predicted from the \emph{feature representations} alone, following prior work in fair representation learning \cite{madras2018learning,zhang2018mitigating}. We find that complex mechanisms by which protected group information can be indirectly predicted could be contributing factors to the fidelity gaps observed.

\subsection{Training Fair Blackbox Models}
Our experiments in previous sections (see Tables~\ref{tab:local} and ~\ref{tab:global}) indicate that fidelity gaps across subgroups occur regardless of the blackbox model's fairness with respect to groundtruth label predictions.
For example, a logistic regression model trained on the \texttt{mimic} dataset is fair with respect to the sex (Table~\ref{tab:blackbox_groundtruth} in Appendix; DP gap of 1\%). However, fidelity gaps are non-zero across sex subgroups with a sparse decision tree global explanation (8.1\%). Strikingly, the gaps in fidelity AUROC often exceed the gaps in AUROC of the blackbox models themselves (e.g., \texttt{mimic} dataset with the Tree model, the difference in classification performance of the blackbox between Male and Female individuals is 3.6\%, while the fidelity gap between subgroups is 8.1\%). However, we do observe larger fidelity gaps in datasets where blackbox models are more accurate but less fair (e.g. \texttt{adult} with an absolute DP gap of 16-17\%).

\textbf{Experiment Setup.} To investigate this further, we train debiased neural network blackbox models for \texttt{lsac} and \texttt{mimic}: both highly-imbalanced by class label, and characterized by the largest and smallest gaps in blackbox model AUROC respectively. Adversarial debiasing following methodology proposed by ~\citet{zhang2018mitigating}\footnote{We use an open-sourced implementation: \url{https://github.com/Trusted-AI/AIF360}} is utilized, wherein an adversary tries to predict the protected group information from classification predictions and labels, while the main classification model (our blackbox model) jointly predicts the primary classification outcome. We use demographic parity as the desired fairness definition.
Our results are shown in Table \ref{tab:fair_blackbox_groundtruth}, which reports the performance of the fair(er) blackbox classifiers.

\textbf{Results.} We debias neural network blackbox models to be fairer, where a model is deemed to be fair if it has an absolute DP gap close to 1\% (9\% and 0.6\% after debiasing for \texttt{lsac} and \texttt{mimic} respectively; improved from 14\% and 2\%). We find that despite fair training, fidelity gaps remain (Table~\ref{fig:Identifiability of groups}, though they are significantly reduced in most cases: the fidelity gap in accuracy decreases from 2\% to 1\% for \texttt{mimic} and 7.4\% to 0.8\% for \texttt{lsac} (Tree)). Note that these results are dependent on our choice of fairness criterion: particularly, for the \texttt{lsac} dataset we find that overall performance is reduced to achieve parity (Table~\ref{tab:fair_blackbox_groundtruth} in Appendix; for absolute DP gap close to 1\%, over 99\% of the blackbox predictions are that the student passes the bar). This indicates that while fair blackboxes could potentially reduce fidelity gaps across subgroups, there might be trade-offs and more potential causes for such gaps. We explore this in Section~\ref{sec:group_pred}.

\subsection{Performance of Predicting Protected Attributes from Feature Representations Alone}
\label{sec:group_pred}
\textbf{Experiment Setup.} One way to achieve group fairness is by removing group information from or debiasing the representations (e.g. with the use of an adversary~\cite{wadsworthachieving,madras2018learning}). Here, we quantify the amount of group information that is present in the data. This is relevant as the absence of group information is a sufficient condition to achieving fairness parity according to standard metrics in fair machine learning.  For example, consider equality of opportunity for the positive class, which can be written as $\hat{Y} \indep G | Y = 1$, where G denotes the protected group and Y is the binary groundtruth label. If we have $X \indep G | Y = 1$, then equality of opportunity is achieved for any form of the classifier $\hat{Y} = f(X)$, including $E$ the explanation model and $B$ the blackbox model. Therefore, if no protected group information is present in the positive examples, then the explanation fidelity would not differ between protected groups for positive examples \citep{madras2018learning}. 
In this experiment, we first compute the accuracy of detecting the protected group information from all datasets (with a cross-validated model). Then, we select features that have zero mutual information with respect to the protected attribute, and only use these in training the blackbox and explanation model. We expect that the performance of predicting group information from these features will be low. Then, we compute the fidelity gaps -- this allows us to answer the question: do fidelity gaps exist when there is low group information in the data?

\begin{figure*}
    \begin{subfigure}[b]{0.48\textwidth}
        \includegraphics[width=0.98\textwidth]{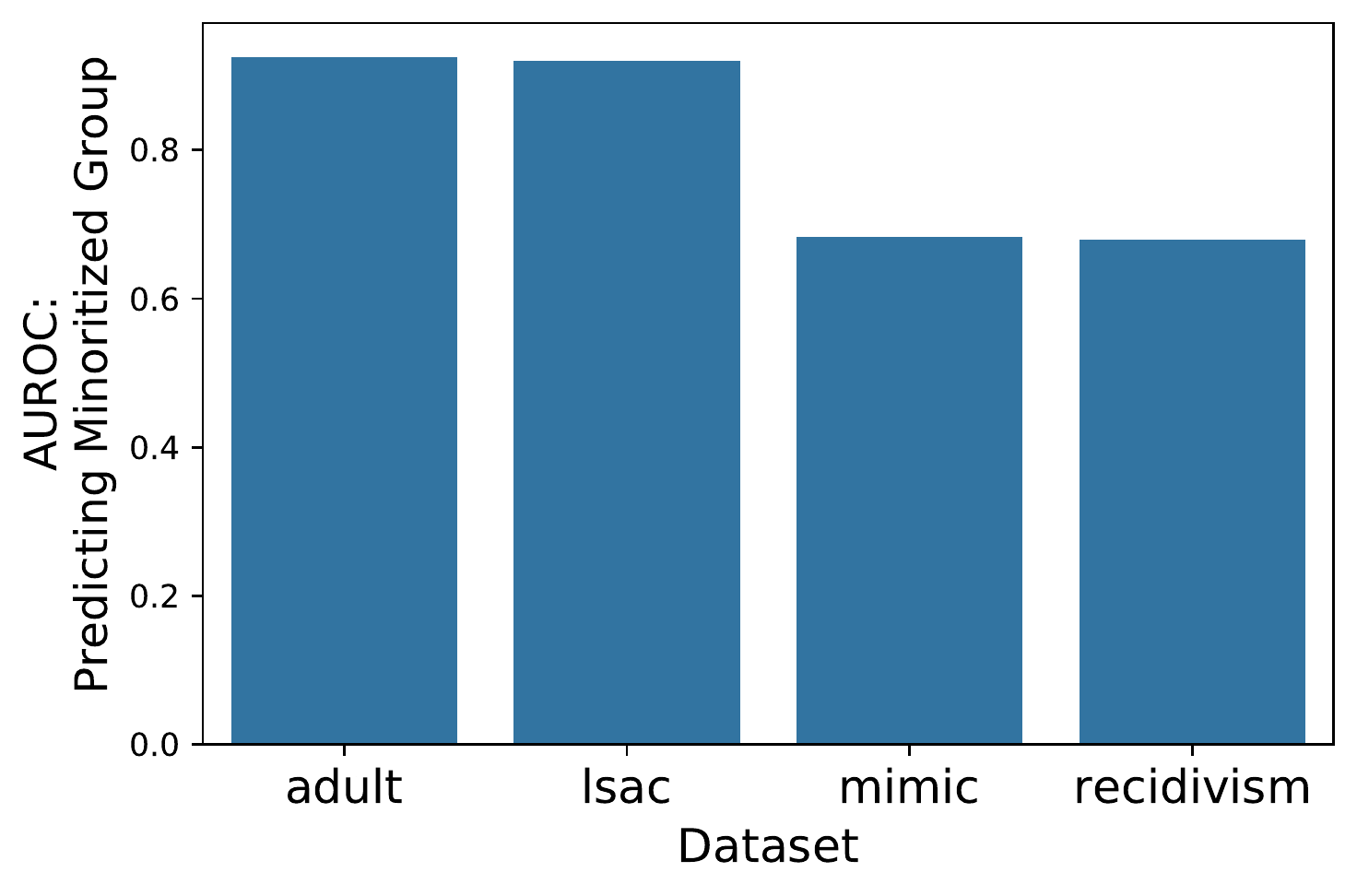}%
        \caption{AUROC in detecting minoritized protected group}
    \end{subfigure}
    \begin{subfigure}[b]{0.4\textwidth}
        \raggedleft
        \begin{tabular}{llccc}
            \toprule
             Dataset & Expl & $Fidelity^{Acc.}$ & $\Delta^\text{group}_{AUROC}$ & $\Delta^\text{group}_{Acc.}$ 
             \\
            \midrule
            
            \multirow{2}{*}{\texttt{lsac}} & GAM &  96.6 &   1.4 &       1.5  \\
            & Tree &   96.9 &  7.0 &       0.8 \\
            
            \midrule                     
            \multirow{2}{*}{\texttt{mimic}} &GAM &  96.2 &    0.6 &       1.0 \\
            & Tree &   94.9 &    4.1 &      1.6 \\
            \bottomrule
        \end{tabular}
        \vspace{8mm} 
        \caption{Fidelity gaps with fair blackbox models}
    \end{subfigure}
    \caption{   
        We find minoritized protected groups can be detected with high AUROC from feature representations alone in (a). As a result, non-zero fidelity gaps persist even when underlying blackboxes are fair as seen in (b). 
    }\label{fig:Identifiability of groups}
\vspace{-5mm}
\end{figure*}

\textbf{Results}: First, we observe that in all cases the prediction AUROC is significantly greater than 0.5 (see Table~\ref{fig:Identifiability of groups}) in identifying the minority group. This indicates that the protected group information -- e.g. if the datapoint belongs to a Male/Female individual -- can be predicted with good performance from the feature representations alone\footnote{This was also observed when conditioned on only positive or negative label classes}.  Since past work has shown that group information might be indirectly constructed and used by explanations~\cite{lakkaraju2020fool}, this is important to consider. Second, we only use features that have zero mutual information with respect to protected attribute labels in the \texttt{lsac} dataset for training  blackbox and explanation models. We see that all models output a single-class prediction which limits our ability to make meaningful conclusions about the impact of group information on this dataset -- note the fidelity gaps are technically zero, though the explanations are trivial. We repeat the same procedure with the \texttt{mimic} dataset by selecting 10 features. The AUROC of predicting protected attribute (sex) from these features is low ($0.54$; also less than $0.57$ for features from positive and negative class). With this representation, we train both NN and LR blackbox models, and GAM/Tree global explanation models. We observe that accuracy-based fidelity gaps ($\Delta_{Acc.}$, $\Delta_{Acc.}^{group}$) decrease to low values not much higher than zero  (to 0-0.6\% with GAM and Tree; full table in Appendix~\ref{tab:group_removed_subset} while blackbox model's AUROC is greater than $0.7$).  This indicates that fidelity gaps decrease when there is less group information in data representations. However, non-zero fidelity gaps in AUROC still persist for Tree-based models (up to 6.6\%). This is due to low prevalence of positive class predictions with the blackbox model on using the reduced data representation ($\approx$3\% positive class at $0.5$ threshold), which has a large impact on AUROC (since it is a ranking-based metric, and sensitive to degree of imbalance). We highlight that more experiments using interpretable, completely group-independent representations (i.e., an AUROC of 0.5 in predicting protected attribute labels) that still have high groundtruth predictive capability are required to accurately quantify the impact of group information on AUROC-based fidelity gaps. We also note that class imbalance -- and varying degrees of class imbalance for data subgroups -- may be an important factor. Our findings indicate that fidelity gaps persist across a range of class-imbalance ratios, but we leave the estimation of the effect of varying degrees of class imbalance (or positive-class prevalence) across subgroups on explanation fairness for future work. 

\section{Simulating the Real-World Impact of Biased Explanations}\label{sec:real_world}

Unfair explanation models can have negative effects on real-world decision making.
To demonstrate this, we conduct a simulation study of ML-assisted law school admissions using the \texttt{lsac} dataset \cite{wightman1998lsac}.
Such systems are already being used in many cases \cite{martinez2021using,pangburn_2019}.
Our results clearly show that worse decisions are made for members of disadvantaged groups when explanations are less fair.



\textbf{Experiment Setup.} To set up our simulation study, we consider an admissions officer that uses a blackbox model that predicts whether a student will pass the bar, though this prediction may be incorrect.
The admissions officer also has an explanation of the model's prediction, which may have low fidelity.
The blackbox model's performance and the explanation fidelity can vary between protected groups---we vary these parameters in this experiment.
We assume that the admissions officer then admits students solely based on their perceived likelihood of the applicant passing the bar, without any knowledge of the applicant's demographics. We assume parameters for the probability that the officer ultimately makes the correct decision. We obtain these parameters from prior user studies assessing the impact of explanations on human+AI decision making accuracy for a different task \cite{bansal2021does}, but believe they serve as reasonable estimates to display the anchoring effect of decisions with explanations observed across a variety of decision-making settings~\cite{poursabzi2021manipulating,bansal2021does}. For further details, please see Appendix \ref{app:sim}.

To simulate the effect of fidelity gaps on decision-making accuracy, we vary the maximum fidelity gaps between the two groups (males and females) and the average from 0\% to 15\%, assuming an average fidelity of 85\% across groups.
We then compute the admissions officer's resulting decision-making accuracy for males and females.
We use sex as the protected attribute of interest for the simulation as both groups pass the bar equally in the dataset, so decision-making accuracy is a fair performance metric.

\begin{figure}
    \centering
      \includegraphics[scale=0.5]{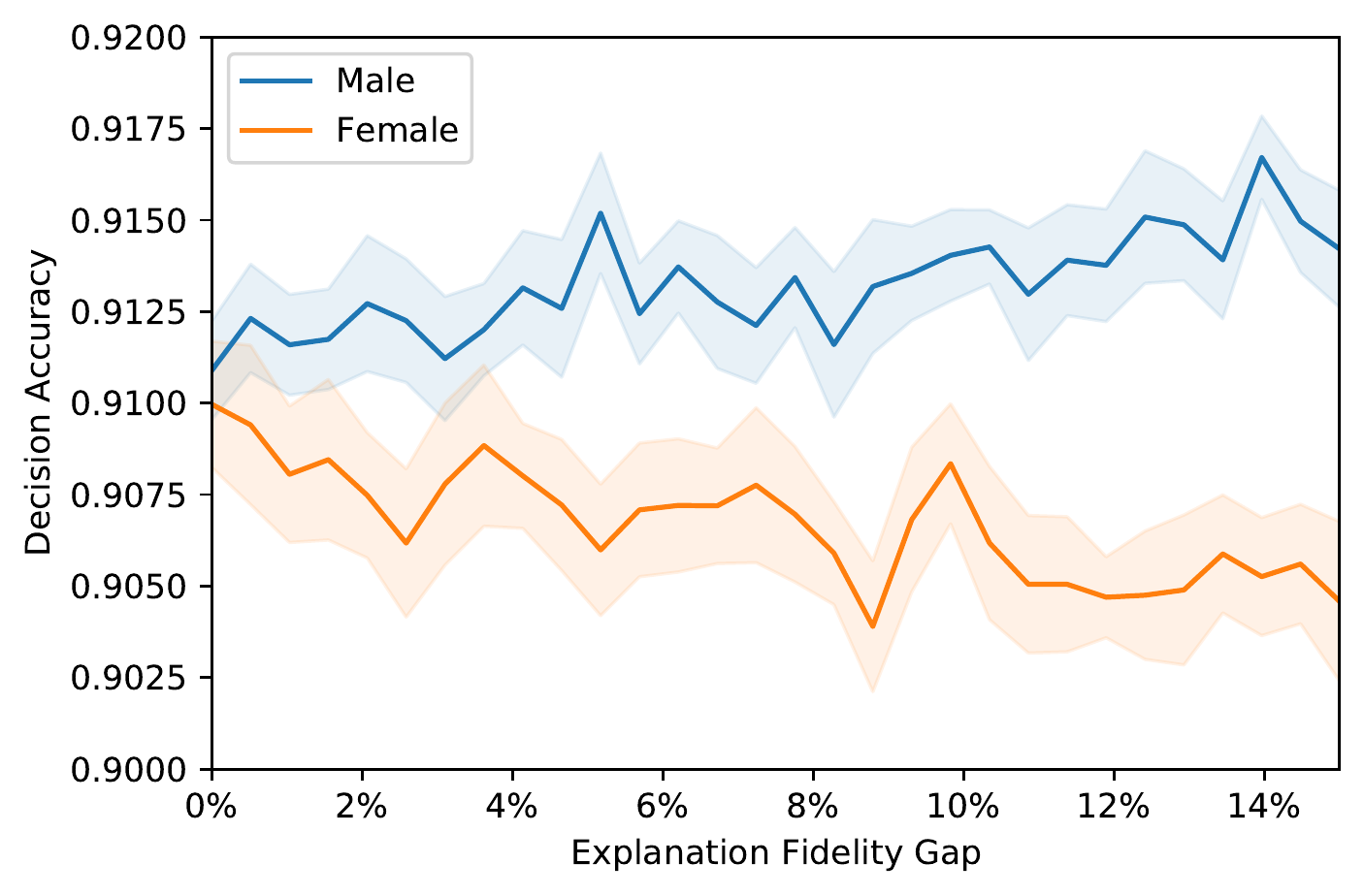}
    \caption{Effect of fidelity gap size on a simulated admissions officer's decision accuracy between males and females using a neural network blackbox. Note that larger fidelity gaps lead to larger decision accuracy gaps between males and females; fidelity gaps could disadvantage different groups in practice. Each line is paired with 95\% confidence intervals across 20 simulations. 
    }\label{fig:expl_gap_sim_nn}
\end{figure}

\textbf{Results.}
As shown in Figure \ref{fig:expl_gap_sim_nn}, we find that larger fidelity gaps lead to larger decision accuracy gap between groups.
So when explanations are less fair, disadvantaged groups may be targeted by worse decisions.
With over 60,000 law school applicants in the U.S. in a typical year \cite{lsac_2018}, over 200 applications would be wrongly admitted/rejected based solely on the fairness of the explanation model, according to this simulation.
Fidelity gaps should therefore be used as fairness metrics for explanation methods: minimizing these gaps leads to fairer decisions.
However, we emphasize that the findings of this simulation are based on some strong assumptions (e.g., reliance on parameters extracted, anchoring effect existence in this admission setting, etc.). Real-world user studies are required to validate these expected findings rigorously across a variety of decision-making setups.

\section{Discussion}

\subsection{Takeaways for ML Practitioners}

\textbf{Analyze subgroup fidelities.} Our results suggest that ML practitioners using post-hoc explainable models to interpret blackbox models should carefully analyse the fidelity of commonly-used explanations for different groups separately. Especially if there is a target subgroup of interest. If a fidelity gap exists, practitioners should carefully consider its source \cite{wick2019unlocking}, and, where possible, take measures to minimize the impacts on downstream decision-making \cite{rajkomar2018ensuring}. We also highlight the importance of carefully choosing the metric for measuring fidelity (e.g., accuracy, AUROC, precision, etc.): different metrics may be affected by properties of the dataset and hence predictions from a blackbox/explanation model (e.g., class imbalance) differently.   

\textbf{Consider the explanation model.} Overall, our findings indicate the existence of fidelity gaps between subgroups is both a \textit{model} and a \textit{data} issue. From Section \ref{sec:fidelity_gaps_exps}, we find that fidelity gaps can vary greatly for the same dataset depending on the explanation model used, and our results in Section \ref{sec:robust_training} show that algorithms that seek to improve worst-case group performance may be a promising direction in reducing fidelity gaps. As such, we recommend careful selection and testing of various explanation models in order to select an equitable model with high overall fidelity. 

In addition, model hyperparameters should also be carefully tuned. For example, in models like LIME, there are several hyperparameters that can affect fidelity gaps, such as the sampling variance (Figure~\ref{fig:sigma_var_lime} in Appendix), the number of perturbations, or number of features in the explanation. Exploring the effect of these hyperparameters on explanation quality and fairness is a promising direction of future work. Lastly, extending prior work on methods for fair supervised ML models~\cite{caton2020fairness,paul2021generalizing,madras2018learning,zhang2018mitigating,liu2021just}, we call for similar approaches to training fair and explainable local and global explanation models which have reduced fidelity gap in addition to high overall fidelity. 

\textbf{Consider the data.} Our results in Section \ref{sec:fair_fidelity_gaps} indicate that fidelity gaps also depend on data representations. Because some feature representations cause smaller fidelity gaps, practitioners should carefully consider the features used to learn both the blackbox and explanation models \cite{grgic2016case}.
As machine learning models can encode historical biases present in the training corpora \cite{doughman2021gender}, it is crucial to consider the source of such potential biases, and, if possible, take actions to correct them by collecting additional data in a fairness-aware way \cite{anahideh2020fair, jo2020lessons}.

\subsection{Implications of Fidelity Gaps}
\textbf{Algorithmic Modifications to Train Fair Explanation Models with Low Fidelity Gaps.} While we benchmark robust training as an attempt to mediate the fidelity gaps across subgroups, we posit that data distribution-aware methods could lead to lower, less significant fidelity gaps. For example, recent work in causal bootstrapping~\cite{imbens2018causal}
shows the potential to reduce confounding biases in ML model training given causal knowledge~\cite{gowda2021pulling}. Similar strategies relying on partial or complete knowledge of the data generation graph~\cite{geirhos2018imagenet} could prove effective in selecting features and training examples to train fair explanation models.

Post-processing solutions to standard explanation model training could also prove effective, similar to recent work in the space of improving worst-case generalization~\cite{menon2020overparameterisation}. However, such solutions need to be appraised carefully to ensure that the resulting models are both \emph{fair} and remain \emph{interpretable} to users~\cite{wang2020pursuit}. 

An interesting follow-up question is whether it is possible to have zero fidelity gaps---perfect worst-case generalization---while retaining good average fidelity under standard training settings. Zero fidelity gaps are possible, of course, when the blackbox and explanation models are identical.
However in more-realistic scenarios, fidelity gaps may simply depend on the data distributions~\cite{menon2020overparameterisation}.
For example, rare subgroups may be more diffidult to approximate, and will naturally have lower fidelity than others~\cite{samadi2018price,hooker2020characterising}.


\textbf{Fidelity Gaps as an Evaluation Metric.} We focus mainly on evaluating the fairness of explanation models using the \textit{fidelity gap} as a metric, assuming that models with smaller fidelity gaps are more desirable. However, recent work in group fairness has found that trying to achieve equal performance for all subgroups tends to worsen welfare for all \cite{hu2020fair, corbett2018measure, zhang2022improving}. Such a fairness/accuracy trade-off is well-documented in the algorithmic fairness literature \cite{kearns2019empirical, zliobaite2015relation}. We posit that there is likely a similar trade-off between the fidelity gap and the overall fidelity of an explanation model. In such cases, motivated by definitions such as minimax Pareto fairness \cite{martinez2020minimax, diana2021minimax}, it may be more appropriate to select explanation models that maximize the fidelity of the worst-case group.

\textbf{Human Implications of Fidelity Gaps.} Explainable ML models form an integral part of sociotechnical systems, given their user-facing nature~\cite{buccinca2020proxy}. Several works have studied the utility of explanations in human--AI joint decision-making~\cite{bansal2021does}. However, the potential failure modes we identify---fidelity gaps leading to worse explanations for some groups---need to be studied further in the context of real human decision-making (in addition to the simulations we conduct). For accurate decision-making in practice, like learning to defer decisions to an expert~\cite{pmlr-v119-mozannar20b}, it is important to communicate clearly and provide end-users with details of performance caveats~\cite{liang2019implicit}.
This requires collaboration between computer scientists and scholars working in the space of computer-mediated communication. A more design-centric approach is required to bridge the gap between researchers and consumers of these models~\cite{chen2022interpretable}.




\section{Conclusion}
In this work, we investigate fairness properties of post-hoc explainability methods.
We ultimately find that significant gaps in performance exist between groups, indicating that some groups receive better explanations than others.
First, we demonstrate experimentally that significant gaps occur in the two main branches of explanation methods using four explainability methods on four common datasets and two blackbox models.
Second, we present a study of robust training methods for improving these gaps.
We find that these methods can improve the fairness of explanation models in some cases.
Third, using a simulation study, we demonstrate that improving explanation fairness could substantially improve decision making accuracy for underserved groups.
Finally, we pose promising directions enhancing post-hoc explainability methods; future work should focus on ensuring explanation quality does not suffer according to group membership while remaining reliable and accurate.


\begin{acks}
Aparna Balagopalan is supported by a grant from the MIT-IBM Watson AI Lab. Haoran Zhang is supported by a grant from the Quanta Research Institute. Dr. Frank Rudzicz  is supported by a CIFAR AI Chair. Dr. Marzyeh Ghassemi is funded in part by Microsoft Research, and a Canadian CIFAR
AI Chair held at the Vector Institute. Resources used
in preparing this research were provided, in part, by
the Province of Ontario, the Government of Canada
through CIFAR, and companies sponsoring the Vector Institute. We would like to thank Hammaad Adam, Bret Nestor, Natalie Dullerud, Vinith Suriyakumar, Nathan Ng, and three anonymous reviewers for their valuable feedback.
\end{acks}

\bibliographystyle{ACM-Reference-Format}
\bibliography{sample-base}

\clearpage
\appendix

\section{Fairness Preservation}
\subsection{Proof}
\label{sec:proof}
\begin{enumerate}
    \item Samples $x_{i}$, each associated with a binary valued sensitive attribute $A_{i}\in\{0,1\}$, and $i\in[1,N]$
    \item Pre-trained blackbox classifier, $B$, which generates predictions $B(x_{i})$ for each point
    \item Local explanation model for each test instance, $E_{i}(x_{i})$ or global explanation model $E(x_{i})$ , trained on tuples $(x^{*}_{i},B(x^{*}_{i}))$. For local explanation models, $(x^{*}_{i},B(x^{*}_{i}))$ are usually neighboring points around instance of interest, generated synthetically or sampling nearest neighbors in the training set.
\end{enumerate}
In all cases, $B$ and $E_{i}$ output probabilities $\in[0,1]$ or binary values.

By definition of explanation models being trained to mimic blackbox models (or be reasonable estimators of $B$ with error minimization of errors such as mean squared error): $$\forall i: E_{i}(x_{i}) =  B(x_{i})+\epsilon_{i},$$ where $\epsilon_{i}$ is the residual error between outputs of blackbox and explanation surrogate model. In the case of local explanation models, $E_{i}$ could be distinct for each point $i$. However, for global models, $E_{i}$ is the same function ($E$) for all test instances.
In the section below, we show how gaps between average and subgroup performance for producing explanations is directly tied to demographic parity metrics of the blackbox model and explanation model(s) in each case.



We want to attain sufficient conditions for DP of the explanation model to reflect that of the blackbox classifier, i.e.,
$$DP_{E(x)}$$ can be used as a proxy for $$DP_{B(x)}$$ for the same $x$, i.e. same test instances.

\begin{equation}
    DP_{B(x)} = \mathbb{E}_{x_{i} : A_{i}=1}[B(x_{i})] - \mathbb{E}_{x_{i} : A_{i}=0}[B(x_{i})]
\end{equation}

\begin{equation}
    DP_{E(x)} = \mathbb{E}_{x_{i} : A_{i}=1}[E_{i}(x_{i})] - \mathbb{E}_{x_{i} : A_{i}=0}[E_{i}(x_{i})]
\end{equation}
\begin{equation}
DP_{E(x)}=\mathbb{E}_{x_i : A_i=1}[B(x_i)+\epsilon_{i}] - \mathbb{E}_{x_i: A_i=0}[B(x_i)+\epsilon_{i}]
\end{equation}
\begin{equation}
DP_{E(x)}=\mathbb{E}_{x_{i} : A_{i}=1}[B(x_{i})] - \mathbb{E}_{x_{i} : A_{i}=0}[B(x_{i})] + \mathbb{E}_{x_i : A_i=1}[\epsilon_{i}] - \mathbb{E}_{x_i : A_i=0}[\epsilon_{i}]
\end{equation}
$$\implies DP_{E(x)}= DP_{B(x)} + Error$$

where $Error$=$\mathbb{E}_{x_i : A_i=1}[\epsilon_{i}] - \mathbb{E}_{x_i: A_i=0}[\epsilon_{i}]$, 
and corresponds to difference in average errors for instances $i$ belonging to each group. $\epsilon_{i}$ directly corresponds to the residual error when minimizing the mean square error or cross-entropy error between continous predictions of $B$ and $E_{i}$ for a test instance $i$.

Additionally, the expansion above holds for  probabilistic fairness definitions~\cite{pleiss2017} when model outputs are probabilistic or constrained to be binary. 

\subsection{Empirical Validation}
\label{sec:empirical_validation}
We empirically validate the theorem on global explanations model (Tree) that utilize standard estimators during explanation model training/optimization. We observe that observations hold for both probabilistic and generalized versions of fairness metrics~\cite{pleiss2017}. This is pertinent because local explanation models are trained to imitate the blackbox prediction probabilities. Note that an underlying assumption (which is validated in practice) is that $E$ and $B$ are both defined for the datapoints considered. We observe that all correlations are significant and consistently $1$. 

\begin{figure*}[h]
\centering
\begin{subfigure}{0.4\textwidth}
  \includegraphics[width=\textwidth]{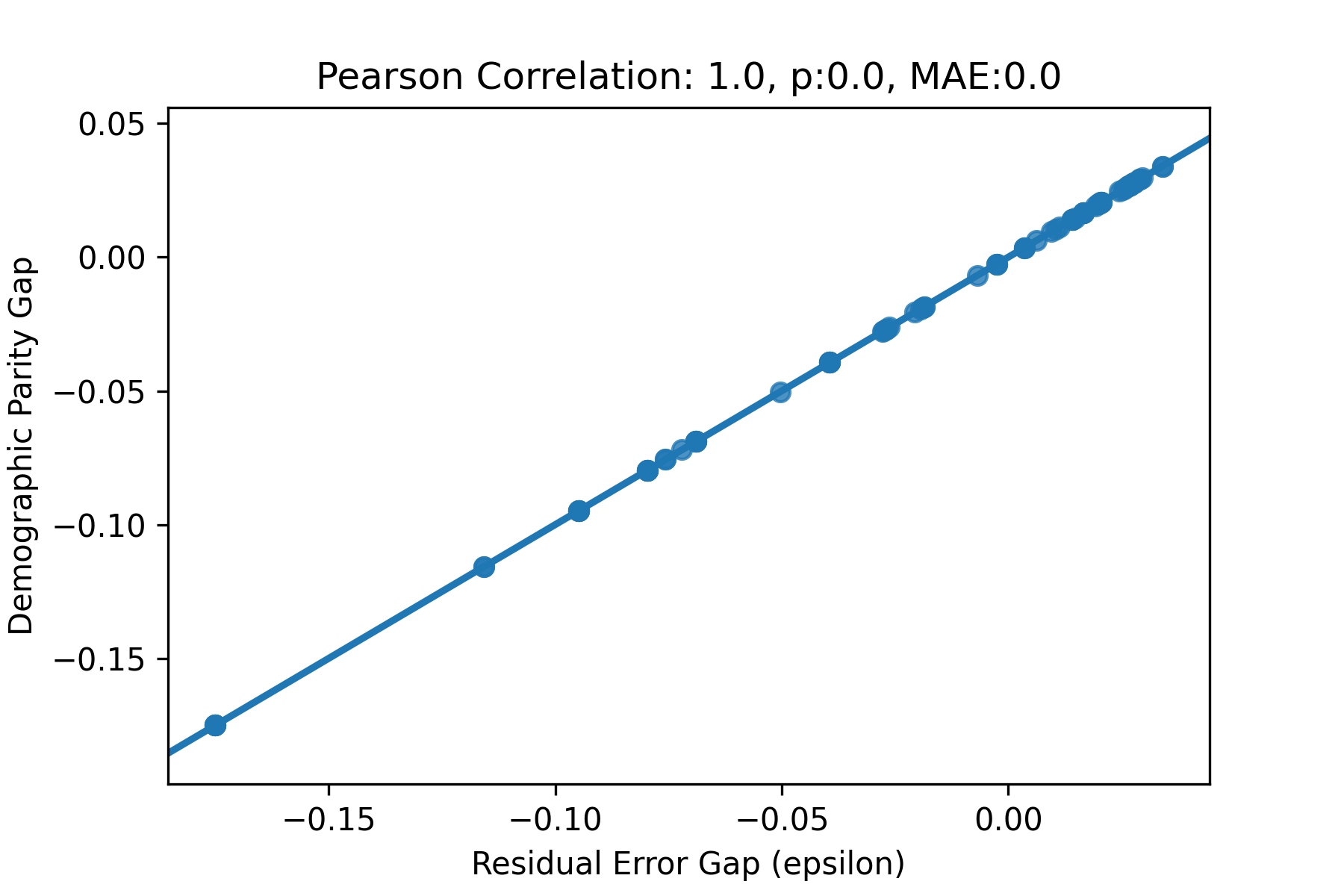}
\caption{Tree + LR blackbox}
\end{subfigure} 
\begin{subfigure}{0.4\textwidth}
  \includegraphics[width=\textwidth]{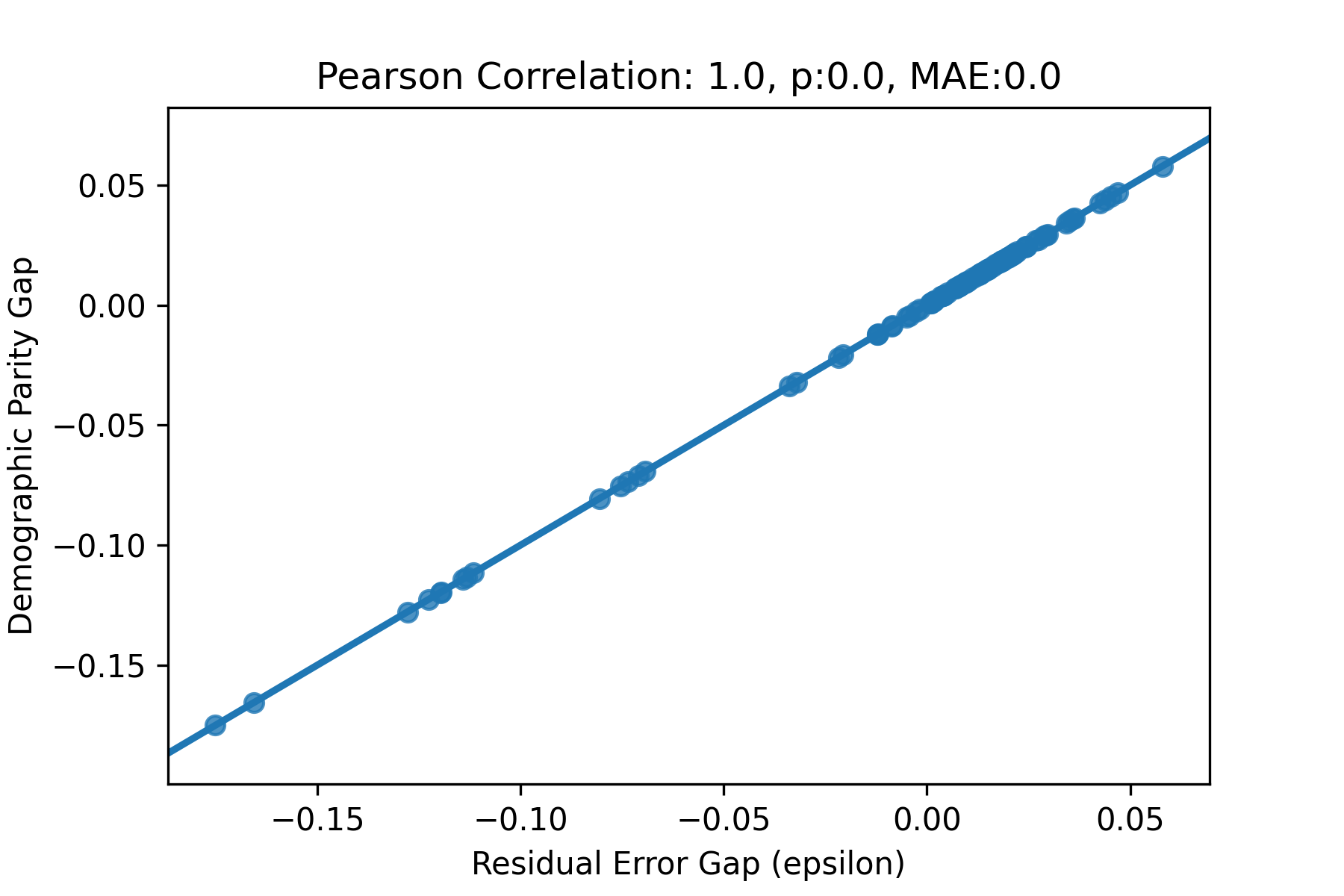}
  \caption{Tree + NN blackbox}
\end{subfigure} \\

  \caption{Comparing gaps in Demographic Parity (DP) of blackbox and explanation model to gaps in residual errors. As per our theorem, these should be the same with 1 and mean absolute error of zero, which is empirically validated. Similar results are observed for all explanation models.}
  \label{fig:empirical_validation}
\end{figure*}

\section{Methods and Experimental Setup}
\label{sec:experimental_setup}

In the sections below, we describe types of post-hoc explanation models used as in our analyses. All experiments are performed for four tabular binary classification datasets (see Section~\ref{sec:datasets}) commonly used in fair machine learning.

\subsection{Post-hoc explanation Models}
\label{sec:explanation_model_descriptions}

We experiment with two local and two global explanation models.

\subsubsection{Local Explanation Models}
Local explanation models explain individual predictions from classifiers by learning an interpretable model locally around each prediction. In our experiments, we consider LIME \cite{ribeiro2016model} and SHAP \cite{lundberg2017unified}, which are popular methods that use linear models to elicit each feature's contribution to the blackbox model's prediction.


Both LIME and SHAP sample points in the neighborhood of query point of interest $\bx$. Then, an auxiliary linear model is trained that approximates the behavior of the blackbox model on the perturbed examples of $\bx'$, while achieving low complexity for interpretability: Both local methods consider a proximity measure $\pi_{\bx'}(\bx)$ between inputs $\bx'$ and $\bx$ to define the neighborhood around a specific datapoint $\bx'$, and $\mathcal{E}$ denote the class of linear models from which we want to choose an explanation model $E$. In both LIME and SHAP, an auxiliary linear model is trained that approximates the behavior of the blackbox model on the perturbed examples of $\bx'$, while achieving low complexity for interpretability:
 $$ \text{argmin}_{E \in \mathcal{E}} L(B, E, \pi_{\bx}) + \Omega(E) $$ where $\Omega(E)$ penalizes the complexity of explanation $E$, and L denotes a loss-function weighted by proximity to $\bx'$.

\paragraph{LIME}
Local Interpretable Model-agnostic Explanations (LIME) \cite{ribeiro2016model} enforces the faithfulness of explanations 
to the blackbox model over a set of examples in the simplified input space. More precisely, LIME does this by fitting a sparse linear model to perturbed samples around query point of interest
$\bm{x}'$. Specifically, we first generate $\mathcal{Z} = (\bm{z}_j, w_j)_{j=1}^{n'}$, where $\bm{z}_j = \bm{x}' + \mathcal{N}(\bm{0}, \sigma^2 \bm{I}_d)$, and $w_j = \mathcal{D}(\bm{z}_j, \bm{x}')$. Here, $\mathcal{D}$ is some distance metric, typically the Euclidean distance.

Taking in $k$, the number of features we wish to explain, LIME then selects $k$ features from $d$, and returns an explanation corresponding to the parameters of a linear model $H: \mathbb{R}^k \rightarrow [0, 1]$. Typically, $H$ is a weighted (by $w$) ridge regression model fit on the sampled datapoints. In this work, we explore the fairness of LIME explanations, varying $\sigma^2$, $k$, and the training procedure of $H$. Note that since LIME implements a regression, the predicted outputs can lie outside $0-1$ range. While the metrics we report do not clip these values, we found that clipping led to the same classification performance. The mean residual error varied with clipping, but is always within 0-0.8\% of the reported value.

\paragraph{SHAP} SHapley Additive exPlanations \cite{lundberg2017unified} uses SHAP values, which are the Shapley values \cite{lundberg2017unified} of a conditional expectation function of the original model.
Similar to LIME, (kernel) SHAP uses a linear model to approximate the blackbox model. Note that the explanation model is a linear function of binary variables $E(z') = \phi_0 + \sum_{i=1}^M \phi_i z'_i$ where $z'=\{0, 1\}^M$ and $M$ is the number of features in the simplified input space. In this approach,  $\{\phi_i\}_{i=1}^{M}$ are Shapley values, which are the only possible solution to the optimization problem that satisfies local accuracy, missingness and consistency~\cite{lundberg2017unified}.

\subsubsection{Global Explanation Models}
\label{sec:global_exp_models_intro}
Global explanation models generate explanations by learning an interpretable model that imitates blackbox model predictions for a set of datapoints. In our experiments, we consider GAM \cite{hastie2017generalized} and a sparse decision tree, which are popular methods that build surrogate models.

\paragraph{Generalized Additive Models (GAM)} 
A generalized additive model (GAM) is a  linear model in which the output variable is modelled as a linear combination of unknown smooth functions of some predictor variables or features. Training involves the inference of these smooth functions. 

For a $d$-dimensonal input $x$,  with predicted variable $y$, $i \in [1,d]$ a GAM can be written as:
$\theta(y) =  \phi_{0}+ \sum_{i=1}^d \phi_i( x_i)$.
where $\theta$ is a link function such as logit or identity.
GAMs are interpretable because the function of each feature on the prediction can be visualized. Following past work~\cite{tan2018distill}, we use GAMs to predict or distill decisions made by blackbox models.

\paragraph{Sparse decision tree (Tree)} We also fit low-depth, sparse, decision trees to predict blackbox decisions. This is because decision trees are simple to understand and to interpret, and can easily be visualised.

There are three primary reasons why these models may approximate a blackbox poorly: (1) disparities in model classes, (2) variance between training datasets for the blackbox and explainability models, and (3) choices of features used for explanation---some methods create explainable surrogate models by using fewer features.

\subsection{Datasets}
\label{sec:datasets}
We conduct experiments using the four tabular binary classification datasets described in Table \ref{tab:datasets}. Each dataset includes labels for each instance's group membership, in this case either their sex or their race. All four datasets are commonly used by prior work on fairness in machine learning \cite{aivodji2021characterizing, suriyakumar2021chasing}. Given the extremely imbalanced nature of the \texttt{lsac} dataset (95\% positive class), we find that even highly-optimized blackbox models (without any fairness criterion) tend to produce less than 1\% negative class predictions. As a result, the AUROC becomes very unstable across consecutive runs. To avoid this, we balance the label-proportions for both classes by randomly upsampling the minority class while training the blackbox model (i.e., 50-50 class balance to train the blackbox model) unless specified otherwise. Proportions of positive-class in groundtruth as well as blackbox model predictions are in Table~\ref{tab:datasets_prevalence} for all datasets.

Following ~\citet{aivodji2021characterizing}, we randomly split each dataset into four subsets: a training set for blackbox models (50\%), a training set for explanation models (30\%), a validation set for explanation models (10\%), and a held-out test set for evaluating both blackbox and explanation models (10\%). In our experiments, we omit the sensitive attribute from all models to allow for closer comparison to previous works.

\subsection{Training Regimes and Hyperparameter Tuning}
\label{sec:training_regimes}

\subsubsection{Training Blackbox Models}
We train two different types of blackbox models on each dataset: Logistic Regression, and a Neural Network.
All models are implemented using the \textit{sklearn} python package~\cite{pedregosa2011scikit}. We tune hyperparameters using a grid-search with five-fold cross validation. We report results for the Logistic Regression (LR) and Neural Network (NN) models in the main text, and report hyperparameter details and performance measures for each in the Appendix. We one-hot encode categorical features, and standardize continuous features to zero-mean and unit variance.
The held-out test sets are used to estimate the accuracy of the blackbox models as well as the generalization of the explanation models. For local explanations, the performance is evaluated only on the query point of interest in each case (and not the synthetic data generated).


\subsubsection{Training Explanation Models}
We use the default settings for LIME~\cite{ribeiro2016model}\footnote{\url{https://github.com/marcotcr/lime}} and SHAP~\cite{lundberg2017unified}\footnote{\url{https://github.com/slundberg/shap}} unless specified otherwise.
For LIME, we use 5000 synthetic data points for each query point. For SHAP, we use 50 random samples. Note that SHAP requires access to a sample of the training set of models, as well as preprocessing steps for datasets with categorical variables. 
For global explanation models, we code categorical features as factors in logistic GAM models (continuous features were fit using splines).
All features are scaled, and numeric features are standardized before fitting global models.
For local models, we consider probabilistic outputs from the blackbox models and for global methods we threshold these predictions at $0.5$ (following ~\citet{aivodji2021characterizing}).


\subsubsection{Choice of Evaluation Metrics} As introduced in Section \ref{sec:metrics}, we measure fidelity gaps between subgroups using two key metrics.
For both metrics (Definitions ~\ref{bg:def:subgroup_avg_gap} and ~\ref{bg:def:subgroup_max_gap}), we must also select a performance metric $L$. We use $L$ to be the AUROC since it is threshold-independent, but also report results using Accuracy as $L$, following past work in explainability~\cite{aivodji2019fairwashing}. Given the imbalanced nature of the datasets considered, AUROC is used for hyperparameter tuning unless specified otherwise.
Note that a non-zero fidelity gap across subgroups measured using both AUROC and accuracy indicates varying degrees of class separability for subgroups in the explanation model. However, the value of the accuracy fidelity gap is also a function of the decision threshold. Here, we select a threshold value of $0.5$ for demonstration purposes, and use the same threshold for all groups. For accurate estimates of overall and subgroup fidelity (gaps), we average across multiple seeds (e.g., prior to performing the $max$ operation for $\Delta_{L}$ computation).

\section{Training Blackbox Models}
\subsection{Hyperparameter Search Space}
We perform 5-fold grid-search hyperparameter tuning to maximize the AUROC of blackbox models in predicting the groundtruth label. We also found that tuning for F1-score macro-averaged at a threshold of $0.5$ led to same hyperparameter settings (i.e., calibrating the blackbox models at a threshold of $0.5$ is justified).
\begin{itemize}
    \item Logistic Regression (LR): We implement this using scikit-learn~\cite{pedregosa2011scikit}, and use the linear solver with L2 regularization. Grid-search hyperparameter tuning is performed for regularization constants between 1e-5 to 1 (over 25 evenly-spaced values).
    \item Neural Network (NN): We implement a neural network with one hidden layer, and vary the number of units from 50, 100, and 200 with grid-search hyperparameter tuning. The relu activation function is used at each hidden
layer, and the network is trained using Adam for up to 200 epochs with other default parameters in ~\citet{pedregosa2011scikit}.
\end{itemize}

\subsection{Performance of Blackbox Models in Predicting Groundtruth}

We ensure that all blackbox models have an average accuracy@chosen threshold of 0.5 greater than 0.68 and an average AUROC greater than 0.7, as observed in Table~\ref{tab:blackbox_groundtruth}. For all datasets blackbox models are well-calibrated with Brier scores~\cite{brier1950verification} ranging from 0.06-0.29 across the subgroups. We observe that several blackbox models are unfair: both in terms of varying performance across groups as well as their Demographic Parity Gaps.

\begin{table*}[ht]
\begin{tabular}{llllrrrrr}
\toprule
Dataset & Blackbox  &  AUROC & Acc. & $\Delta_{AUROC}$ & $\Delta_{Acc.}$ & $\Delta^{group}_{AUROC}$ & $\Delta^{group}_{Acc.}$ & $|DP|$ \\
\midrule
adult &             lr &              90.1\% ± 0.0\% &  85.3\% ± 0.0\% &   2.7\% ± 0.0\% &      3.8\% ± 0.0\% &    6.1\% ± 0.0\% &      11.1\% ± 0.0\% &         17.5\% ± 0.0\% \\
 &             nn &              90.8\% ± 0.2\% &  85.8\% ± 0.1\% &   2.4\% ± 0.0\% &      3.8\% ± 0.1\% &    5.3\% ± 0.1\% &      11.3\% ± 0.4\% &         16.1\% ± 0.3\% \\
lsac &             lr &             88.8\% ± 0.0\% &  78.5\% ± 0.2\% &  23.0\% ± 0.1\% &     29.0\% ± 0.5\% &   14.2\% ± 0.2\% &      16.4\% ± 0.3\% &         25.9\% ± 0.2\% \\
 &  nn &   82.9\% ± 1.3\% &  88.8\% ± 1.1\% &  21.8\% ± 1.6\% &     17.7\% ± 3.0\% &   14.0\% ± 3.5\% &      11.7\% ± 2.0\% &         13.7\% ± 0.8\% \\
mimic &             lr &   
80.6\% ± 0.0\% &  89.4\% ± 0.0\% &   2.0\% ± 0.0\% &      0.9\% ± 0.0\% &    3.6\% ± 0.0\% &       1.6\% ± 0.0\% &          1.4\% ± 0.0\% \\
 &             nn &              78.4\% ± 0.7\% &  88.1\% ± 0.3\% &   1.1\% ± 1.0\% &      1.3\% ± 0.4\% &    2.0\% ± 1.7\% &       2.3\% ± 0.7\% &          2.0\% ± 0.7\% \\
recidivism &    lr &        72.5\% ± 0.0\% &  68.5\% ± 0.0\% &   2.1\% ± 0.0\% &      0.9\% ± 0.0\% &    3.5\% ± 0.0\% &       1.5\% ± 0.0\% &         22.1\% ± 0.0\% \\
 &             nn &            71.8\% ± 0.1\% &  67.5\% ± 0.3\% &   1.8\% ± 0.1\% &      1.5\% ± 0.3\% &    3.0\% ± 0.1\% &       2.4\% ± 0.4\% &         21.8\% ± 0.2\% \\

\bottomrule
\end{tabular}
\caption{Blackbox Model Performance in Predicting Groundtruth\label{tab:blackbox_groundtruth}}
\end{table*} 

\subsection{Prevalence of Positive Class}
The datasets we study have varying degrees of class imbalance in groundtruth as well as in predictions of trained blackbox models (at a calibrated decision threshold of $0.5$; see Table~\ref{tab:datasets_prevalence}). We see that fidelity gaps are observed across these different prevalence rates, though class imbalance could have an impact on the sensitivity of the \emph{metrics}. This underscores the importance of choosing the evaluation metric based on deployment context carefully. Controlled experiments where we oversampled the minority label class led to similar degrees of fidelity gaps for all datasets (see Table~\ref{tab:balanced_labels_both} in Appendix).

\begin{table}[]
\begin{tabular}{llccc}
\toprule
\textbf{Dataset} & \textbf{Outcome Variable} & \textbf{Majority Class\%} & \textbf{Blackbox} & \textbf{Predicted Majority Class\%}   \\ \midrule
\texttt{adult} \cite{Dua:2019} & Income $>$ 50K    & 78 & lr & 81\\
& & & nn & 80\\
\midrule
\texttt{lsac} \cite{wightman1998lsac} & Student passes the bar    & 95 & lr & 75\\
& & & nn & 88\\
\midrule
\texttt{mimic} \cite{harutyunyan2019multitask} & Patient dies in ICU  & 88 & lr & 96\\
& & & nn & 92\\
\midrule
\texttt{recidivism} \cite{propublica2019} & Defendant re-offends       & 55 & lr & 65\\ 
& & & nn & 62\\
\bottomrule
\end{tabular}
\caption{Binary classification datasets used in our experiments. The prevalence of majority class (or degree of imbalance) on the test set is detailed above.  \label{tab:datasets_prevalence}}
\end{table}

\section{FIDELITY GAPS EXIST IN LOCAL AND GLOBAL EXPLANATION MODELS}
\label{sec:full_gaps_tables}
We observe significant gaps in fidelity between data subgroups. The largest gaps are generally observed for the \texttt{lsac} dataset, and this could be because there are annotations available for more sensitive subgroups (5 race subgroups). 
\begin{table*}[ht]
\begin{tabular}{llllrrrr}
\toprule
Dataset & Blackbox & Expl & $Fidelity^{AUROC}$ & $\Delta_{AUROC}$ & $\Delta_{Acc.}$ & $\Delta^{group}_{AUROC}$ & $\Delta^{group}_{Acc.}$ \\
\midrule
\multirow{2}{*}{\texttt{adult}} &             lr &              LIME &   99.9\% ± 0.0\% &   0.0\% ± 0.0\% &      0.8\% ± 0.0\% &    0.1\% ± 0.0\% &       2.4\% ± 0.1\%  \\
 &             nn &              LIME &   95.7\% ± 1.2\% &   1.1\% ± 0.5\% &      6.9\% ± 0.7\% &    3.0\% ± 1.2\% &      20.6\% ± 2.0\% \\
\midrule
\multirow{2}{*}{\texttt{lsac}} &             lr &              LIME &   100.0\% ± 0.0\% &   0.0\% ± 0.0\% &      2.0\% ± 1.0\% &    0.0\% ± 0.0\% &       1.5\% ± 0.5\% \\
 &             nn &              LIME &   93.5\% ± 0.5\% &   9.8\% ± 3.2\% &     21.4\% ± 4.4\% &    6.6\% ± 1.2\% &      12.2\% ± 2.2\% \\
 \midrule
\multirow{2}{*}{\texttt{mimic}} &             lr &              LIME &  84.2\% ± 1.3\% &   1.1\% ± 1.3\% &      0.4\% ± 0.6\% &    3.0\% ± 1.8\% &       1.1\% ± 0.3\% \\ 
 &             nn &              LIME &  92.6\% ± 3.7\% &   0.9\% ± 0.8\% &      0.8\% ± 0.4\% &    1.7\% ± 1.5\% &       1.4\% ± 0.7\% \\ 
\midrule
\multirow{2}{*}{\texttt{recidivism}} &             lr &              LIME &  100.0\% ± 0.0\% &   0.0\% ± 0.0\% &      0.1\% ± 0.1\% &    0.0\% ± 0.0\% &       0.3\% ± 0.2\% \\
 &             nn &              LIME &   99.0\% ± 0.2\% &   0.3\% ± 0.1\% &      0.9\% ± 0.3\% &    0.7\% ± 0.3\% &       2.4\% ± 0.7\% \\
\bottomrule
\end{tabular}
\caption{Gaps in Local Explanation Models; all models have fidelity greater than 84\%\label{tab:full_gaps_local}}
\end{table*}

\begin{table*}[ht]
\begin{tabular}{llllrrrr}
\toprule
Dataset & Blackbox & Expl & $Fidelity^{AUROC}$ & $\Delta_{AUROC}$ & $\Delta_{Acc.}$ & $\Delta^{group}_{AUROC}$ & $\Delta^{group}_{Acc.}$ \\
\midrule
\multirow{4}{*}{\texttt{adult}} &             lr &              GAM &  100.0\% ± 0.0\% &   0.0\% ± 0.0\% &      0.1\% ± 0.0\% &    0.0\% ± 0.0\% &       0.3\% ± 0.0\% \\
 &             lr &             Tree &  96.1\% ± 0.2\% &   1.3\% ± 0.0\% &      1.5\% ± 0.1\% &    2.9\% ± 0.4\% &       4.5\% ± 0.2\% \\
 &             nn &              GAM &  98.8\% ± 0.2\% &   0.2\% ± 0.1\% &      0.8\% ± 0.2\% &    0.5\% ± 0.3\% &       2.4\% ± 0.5\% \\
 &             nn &              Tree &  95.9\% ± 0.3\% &   0.8\% ± 0.6\% &      1.1\% ± 0.1\% &    0.6\% ± 0.4\% &       3.4\% ± 0.2\% \\
\midrule
\multirow{4}{*}{\texttt{lsac}} &             lr &    GAM &          100.0\% ± 0.0\% &   0.0\% ± 0.0\% &      0.9\% ± 0.9\% &    0.0\% ± 0.0\% &       0.6\% ± 0.4\% \\
 &             lr &             Tree & 98.3\% ± 0.1\% &   2.0\% ± 0.4\% &      3.7\% ± 3.1\% &    1.1\% ± 0.4\% &       2.8\% ± 0.7\% \\
 &             nn &              GAM &  93.2\% ± 0.6\% &   8.5\% ± 1.8\% &     13.5\% ± 0.9\% &    5.2\% ± 1.2\% &       7.3\% ± 1.0\% \\
 &             nn &              Tree &  91.1\% ± 0.8\% &   5.0\% ± 1.9\% &     11.5\% ± 2.7\% &    5.8\% ± 2.1\% &       7.4\% ± 1.2\% \\
 \midrule
\multirow{4}{*}{\texttt{mimic}} &             lr &              GAM &  99.5\% ± 0.1\% &   0.2\% ± 0.1\% &      0.5\% ± 0.1\% &    0.4\% ± 0.1\% &       0.9\% ± 0.1\% \\
 &             lr &             Tree &  84.6\% ± 0.4\% &   3.7\% ± 0.4\% &      0.6\% ± 0.0\% &    8.1\% ± 0.8\% &       1.2\% ± 0.1\% \\
 &             nn &             GAM & 90.6\% ± 4.4\% &   0.4\% ± 1.1\% &      1.2\% ± 0.3\% &    1.8\% ± 1.2\% &       2.2\% ± 0.6\% \\
 &             nn &              Tree &  73.4\% ± 4.0\% &   1.4\% ± 0.7\% &      1.1\% ± 0.5\% &    3.0\% ± 1.5\% &       2.0\% ± 0.9\% \\
\midrule
\multirow{4}{*}{\texttt{recidivism}} &             lr &             GAM &  99.9\% ± 0.0\% &   0.0\% ± 0.0\% &      0.1\% ± 0.0\% &    0.1\% ± 0.0\% &       0.3\% ± 0.0\% \\
 &             lr &              Tree &  98.6\% ± 0.0\% &   0.4\% ± 0.0\% &      0.0\% ± 0.0\% &    0.4\% ± 0.0\% &       0.1\% ± 0.0\% \\
 &             nn &              GAM &  99.3\% ± 0.1\% &   0.2\% ± 0.1\% &      0.2\% ± 0.2\% &    0.4\% ± 0.2\% &       0.6\% ± 0.6\% \\
 &             nn &              Tree &  98.9\% ± 0.6\% &   0.4\% ± 0.3\% &      0.9\% ± 0.3\% &    1.0\% ± 0.9\% &       2.3\% ± 0.7\% \\
\bottomrule
\end{tabular}
\caption{Gaps in Global Explanation Models; all models have fidelity greater than 72\%\label{tab:full_gaps_global}}
\end{table*}

We also verify that fidelity gaps exist for the \texttt{lsac} dataset even without label-balanced blackbox models (as described Section~\ref{sec:datasets}). These results are summarized in Table~\ref{tab:lsac_unsampled_version}.

\begin{table*}[ht]
\begin{tabular}{llrr}
\toprule
Blackbox & Expl &  $\Delta^{group}_{AUROC}$ & $\Delta^{group}_{Acc.}$\\
\midrule
\multirow{3}{*}{lr} &  LIME & 0.1\% ± 0.0\% & 4.6\% ± 0.0\%\\
& GAM & 0.1\% ± 0.0\% &  2.6\% ± 0.0\%\\
& Tree &  3.9\% ± 0.0\% &  2.7\% ± 0.0\%  \\
\midrule
\multirow{3}{*}{nn} &  LIME & 2.2\% ± 1.5\% &2.9\% ± 0.8\%\\
& GAM &  4.9\% ± 5.4\% &  3.0\% ± 0.8\%  \\
 &             Tree & 21.7\% ± 5.2\% & 3.9\% ± 1.2\%\\
\bottomrule
\end{tabular}
\caption{Gaps in Explanation Models: \texttt{lsac} dataset when the blackbox model is trained without label-balancing described in Section~\ref{sec:datasets}.\label{tab:lsac_unsampled_version}}
\end{table*}

\section{Gaps Exist in Blackbox Models Trained on Balanced Train Sets}
We observe that even with training sets balanced by protected group labels -- by randomly oversampling the minority group -- significant, non-zero fidelity gaps still persist (see Table~\ref{tab:balanced_local} and Table~\ref{tab:balanced_global}). 
\begin{table*}[ht]
\begin{tabular}{llllrrrr}
\toprule
Dataset & Blackbox & Expl & $Fidelity^{AUROC}$ & $\Delta_{AUROC}$ & $\Delta_{Acc.}$ & $\Delta^{group}_{AUROC}$ & $\Delta^{group}_{Acc.}$ \\
\midrule
\multirow{4}{*}{\texttt{adult}} &             lr &              LIME &   99.9\% ± 0.0\% &   0.0\% ± 0.0\% &      0.8\% ± 0.1\% &    0.1\% ± 0.0\% &       2.3\% ± 0.2\% \\
 &             lr &              SHAP &  100.0\% ± 0.0\% &   0.0\% ± 0.0\% &      0.0\% ± 0.0\% &    0.0\% ± 0.0\% &       0.0\% ± 0.0\% \\
 &             nn &              LIME &   92.1\% ± 1.9\% &   1.6\% ± 0.2\% &      6.4\% ± 1.4\% &    4.6\% ± 0.7\% &      19.1\% ± 4.2\% \\
 &             nn &              SHAP &  100.0\% ± 0.0\% &   0.0\% ± 0.0\% &      0.0\% ± 0.0\% &    0.0\% ± 0.0\% &       0.0\% ± 0.0\% \\
\midrule
\multirow{4}{*}{\texttt{lsac}} &             lr &              LIME &   100.0\% ± 0.0\% &   0.2\% ± 0.2\% &      5.9\% ± 1.3\% &    0.2\% ± 0.1\% &       3.4\% ± 0.8\% \\
 &             lr &              SHAP &  100.0\% ± 0.0\% &   0.0\% ± 0.0\% &      0.0\% ± 0.0\% &    0.0\% ± 0.0\% &       0.0\% ± 0.0\% \\
 &             nn &              LIME &   88.9\% ± 0.9\% &  11.5\% ± 3.3\% &     12.6\% ± 1.6\% &    8.6\% ± 1.7\% &       6.7\% ± 1.1\% \\
 &             nn &              SHAP &  100.0\% ± 0.0\% &   0.0\% ± 0.0\% &      0.0\% ± 0.0\% &    0.0\% ± 0.0\% &       0.0\% ± 0.0\% \\
 \midrule
\multirow{4}{*}{\texttt{mimic}} &             lr &              LIME &  82.7\% ± 0.6\% &   1.1\% ± 1.1\% &      0.4\% ± 0.6\% &    2.2\% ± 2.2\% &       1.0\% ± 1.0\% \\ 
 &             lr &              SHAP &  100.0\% ± 0.0\% &   0.0\% ± 0.0\% &      0.0\% ± 0.0\% &    0.0\% ± 0.0\% &       0.0\% ± 0.0\% \\
 &             nn &              LIME & 93.5\% ± 1.4\% &   0.8\% ± 0.7\% &      1.2\% ± 0.6\% &    1.6\% ± 1.4\% &       2.2\% ± 1.0\% \\
 &             nn &              SHAP &  100.0\% ± 0.0\% &   0.0\% ± 0.0\% &      0.0\% ± 0.0\% &    0.0\% ± 0.0\% &       0.0\% ± 0.0\% \\
\midrule
\multirow{4}{*}{\texttt{recidivism}} &             lr &              LIME &  100.0\% ± 0.0\% &   0.0\% ± 0.0\% &      0.0\% ± 0.1\% &    0.0\% ± 0.0\% &       0.1\% ± 0.2\% \\
 &             lr &              SHAP &  100.0\% ± 0.0\% &   0.0\% ± 0.0\% &      0.0\% ± 0.0\% &    0.0\% ± 0.0\% &       0.0\% ± 0.0\% \\
 &             nn &              LIME &   98.2\% ± 0.5\% &   0.5\% ± 0.1\% &      1.7\% ± 0.7\% &    1.0\% ± 0.3\% &       4.6\% ± 1.9\% \\
 &             nn &              SHAP &  100.0\% ± 0.0\% &   0.0\% ± 0.0\% &      0.0\% ± 0.0\% &    0.0\% ± 0.0\% &       0.0\% ± 0.0\% 
\\
\bottomrule
\end{tabular}
\caption{Gaps in Local Explanation Models when Blackbox Models are Trained on Balanced Training Sets\label{tab:balanced_local}}
\end{table*}

\begin{table*}[ht]
\begin{tabular}{llllrrrr}
\toprule
Dataset & Blackbox & Expl & $Fidelity^{AUROC}$ & $\Delta_{AUROC}$ & $\Delta_{Acc.}$ & $\Delta^{group}_{AUROC}$ & $\Delta^{group}_{Acc.}$ \\
\midrule
\multirow{4}{*}{\texttt{adult}} &             lr &              GAM & 100.0\% ± 0.0\% &   0.0\% ± 0.0\% &      0.0\% ± 0.0\% &    0.0\% ± 0.0\% &       0.1\% ± 0.1\% \\
 &             lr &              Tree &  96.2\% ± 0.3\% &   1.2\% ± 0.1\% &      1.4\% ± 0.0\% &    2.8\% ± 1.0\% &       4.2\% ± 0.1\% \\
 &             nn &              GAM & 97.8\% ± 0.7\% &   0.3\% ± 0.2\% &      1.2\% ± 0.3\% &    0.4\% ± 0.4\% &       3.6\% ± 0.9\% \\
 &             nn &              Tree & 95.1\% ± 0.7\% &   0.8\% ± 0.4\% &      1.7\% ± 0.6\% &    1.0\% ± 1.2\% &       5.0\% ± 1.7\% \\
\midrule
\multirow{4}{*}{\texttt{lsac}} &             lr &              GAM &  100.0\% ± 0.0\% &   0.1\% ± 0.1\% &      1.4\% ± 1.1\% &    0.0\% ± 0.0\% &       0.9\% ± 0.7\% \\
 &             lr &              Tree & 98.2\% ± 0.4\% &   4.3\% ± 1.6\% &      7.3\% ± 2.8\% &    2.7\% ± 0.6\% &       3.9\% ± 1.4\% \\
 &             nn &              GAM &  89.5\% ± 1.3\% &  12.7\% ± 3.1\% &     14.4\% ± 2.6\% &    9.4\% ± 2.7\% &       8.5\% ± 2.2\% \\
 &             nn &              Tree &  85.2\% ± 2.5\% &  15.4\% ± 2.7\% &     17.4\% ± 1.9\% &   10.8\% ± 2.7\% &       9.1\% ± 0.5\% \\
 \midrule
\multirow{4}{*}{\texttt{mimic}} &             lr &              GAM & 99.7\% ± 0.1\% &   0.1\% ± 0.0\% &      0.4\% ± 0.2\% &    0.2\% ± 0.1\% &       0.7\% ± 0.3\% \\
 &             lr &              Tree &  87.2\% ± 1.0\% &   2.8\% ± 0.3\% &      0.6\% ± 0.1\% &    5.8\% ± 0.7\% &       1.2\% ± 0.1\% \\
 &             nn &       GAM &  90.7\% ± 1.9\% &   0.4\% ± 0.8\% &      1.1\% ± 0.4\% &    1.0\% ± 1.2\% &       1.9\% ± 0.7\% \\
 &             nn &              Tree &  72.7\% ± 3.5\% &   0.0\% ± 1.7\% &      1.1\% ± 0.8\% &    2.4\% ± 2.3\% &       1.9\% ± 1.5\% \\
\midrule
\multirow{4}{*}{\texttt{recidivism}} &             lr &              GAM &  99.8\% ± 0.2\% &   0.1\% ± 0.1\% &      0.2\% ± 0.1\% &    0.3\% ± 0.3\% &       0.5\% ± 0.2\% \\
 &             lr   &  Tree &  97.9\% ± 0.5\% &   0.1\% ± 0.2\% &      0.5\% ± 0.5\% &    0.8\% ± 0.9\% &       1.4\% ± 1.1\% \\
 &             nn &              GAM &  99.3\% ± 0.1\% &   0.2\% ± 0.1\% &      0.5\% ± 0.6\% &    0.5\% ± 0.2\% &       1.8\% ± 1.1\% \\
 &             nn &              Tree &  98.6\% ± 0.6\% &   0.3\% ± 0.3\% &      0.6\% ± 0.1\% &    1.0\% ± 0.8\% &       1.7\% ± 0.4\% \\
\bottomrule
\end{tabular}
\caption{Gaps in Global Explanation Models when Blackbox Models are Trained on Balanced Training Sets\label{tab:balanced_global}}
\end{table*}

Similarly, we observe that significant fidelity gaps occur even on oversampling labels during training both blackbox and explanation models. We show the impact of label balancing on during training for \texttt{recidivism} and \texttt{lsac} on $\Delta^{group}_{Acc}$ for two global models below to illustrate this (Table~\ref{tab:balanced_labels_both}). We note that class imbalance -- and varying degrees of class imbalance for data subgroups -- may be an important factor that affects the model training as well as metric computation. We leave the estimation of the effect of balancing degrees of class imbalance as well as balancing of data subgroups on explanation fairness for future work -- our preliminary results indicates that fidelity gaps still might persist with this. 

\begin{table*}
\begin{tabular}{llccc}
            \toprule
             Dataset & Expl & $Fidelity^{Acc.}$ & $\Delta^\text{group}_{AUROC}$ & $\Delta^\text{group}_{Acc.}$ 
             \\
            \midrule
            
            \multirow{2}{*}{\texttt{lsac}} & GAM & 92.7\% ± 0.7\% &    6.7\% ± 1.5\% &       9.5\% ± 1.4\% \\
            & Tree &   90.5\% ± 1.3\% &    5.8\% ± 2.4\% &       9.6\% ± 2.2\% \\
            
            \midrule                     
            \multirow{2}{*}{\texttt{recidivism}} & GAM &  99.2\% ± 0.4\% &    0.5\% ± 0.2\% &       1.5\% ± 0.9\% \\
            & Tree &   98.9\% ± 0.4\% &    0.6\% ± 0.5\% &       1.1\% ± 0.5\% \\
            \bottomrule
        \end{tabular}
        \vspace{8mm} 
        \caption{Fidelity gaps with oversampling class labels (data subgroups still imbalanced) for both blackbox and explanation models.\label{tab:balanced_labels_both}}
\end{table*}

\section{Gaps in Fidelity with Varying Features}
\label{sec:varying_features_plots_all}
\begin{figure}[ht!]
    \centering
    \begin{subfigure}{.49\textwidth}
        \includegraphics[width=1\linewidth]{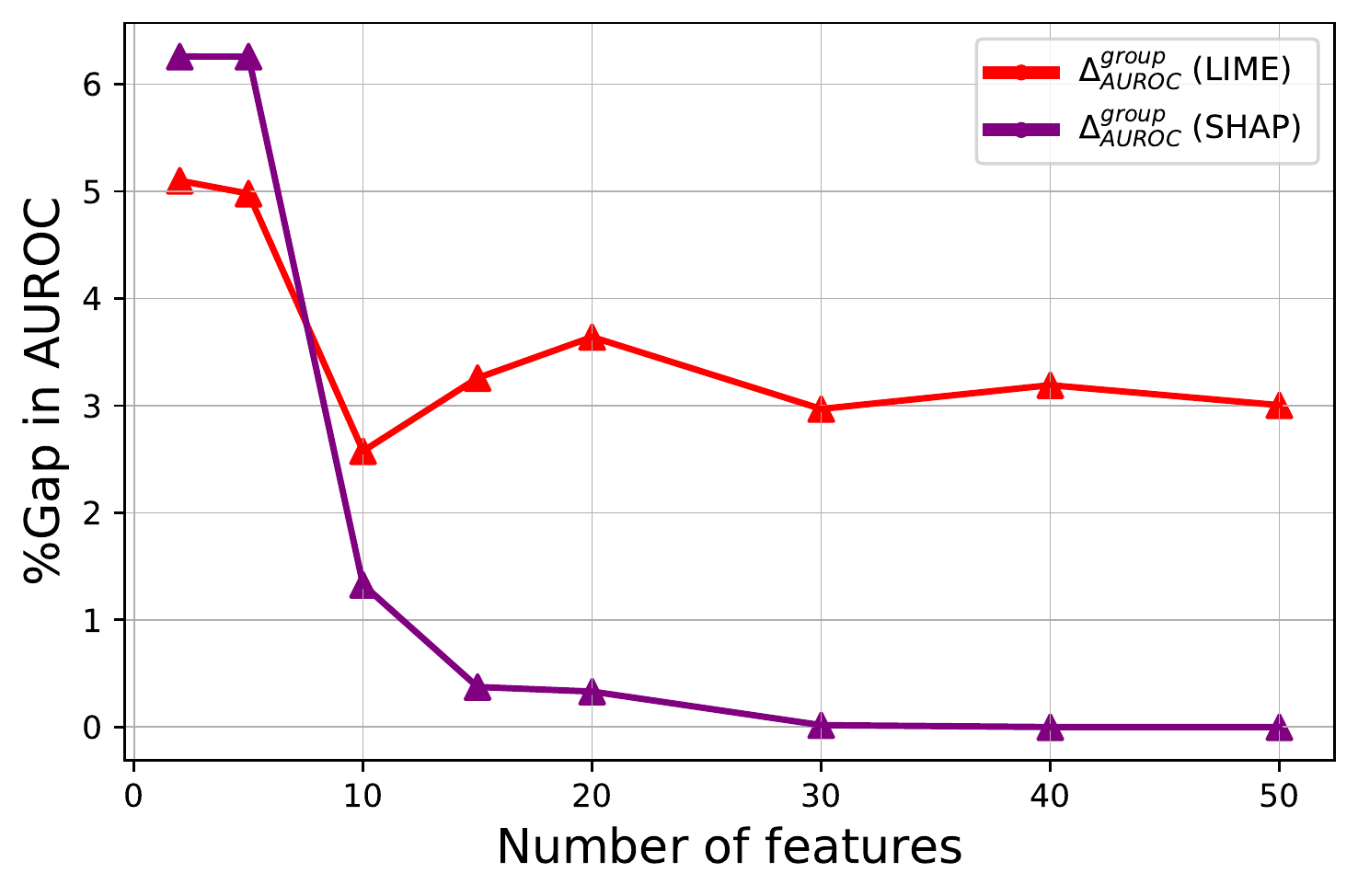}
         \caption{\texttt{adult} + NN + AUROC-fidelity gap for local explanations}
    \end{subfigure}
    \begin{subfigure}{.49\textwidth}
        \includegraphics[width=1\linewidth]{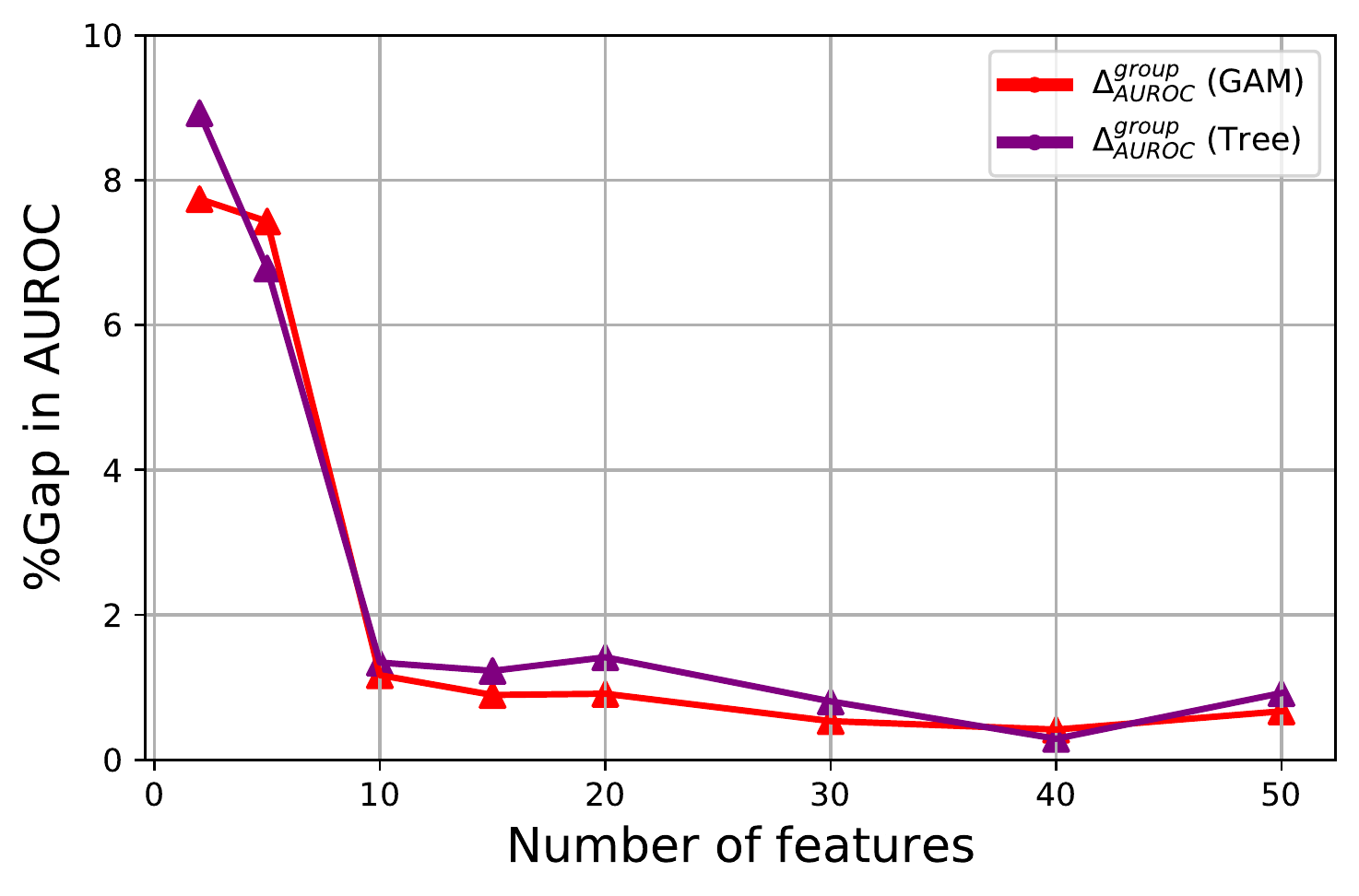}
         \caption{\texttt{adult} + NN + AUROC-fidelity gap for global explanations}
    \end{subfigure}
    \caption{Similar to the figure in the main text, we observe larger fidelity gaps across subgroups with sparser models, i.e., fewer features in local explanation model in AUROC fidelity gaps. The plots shown above are for the \emph{adult} dataset. \label{fig:auroc_gaps_adult}}
\end{figure}

\begin{figure}[ht!]
    \centering
    \begin{subfigure}{.49\textwidth}
        \includegraphics[width=1\linewidth]{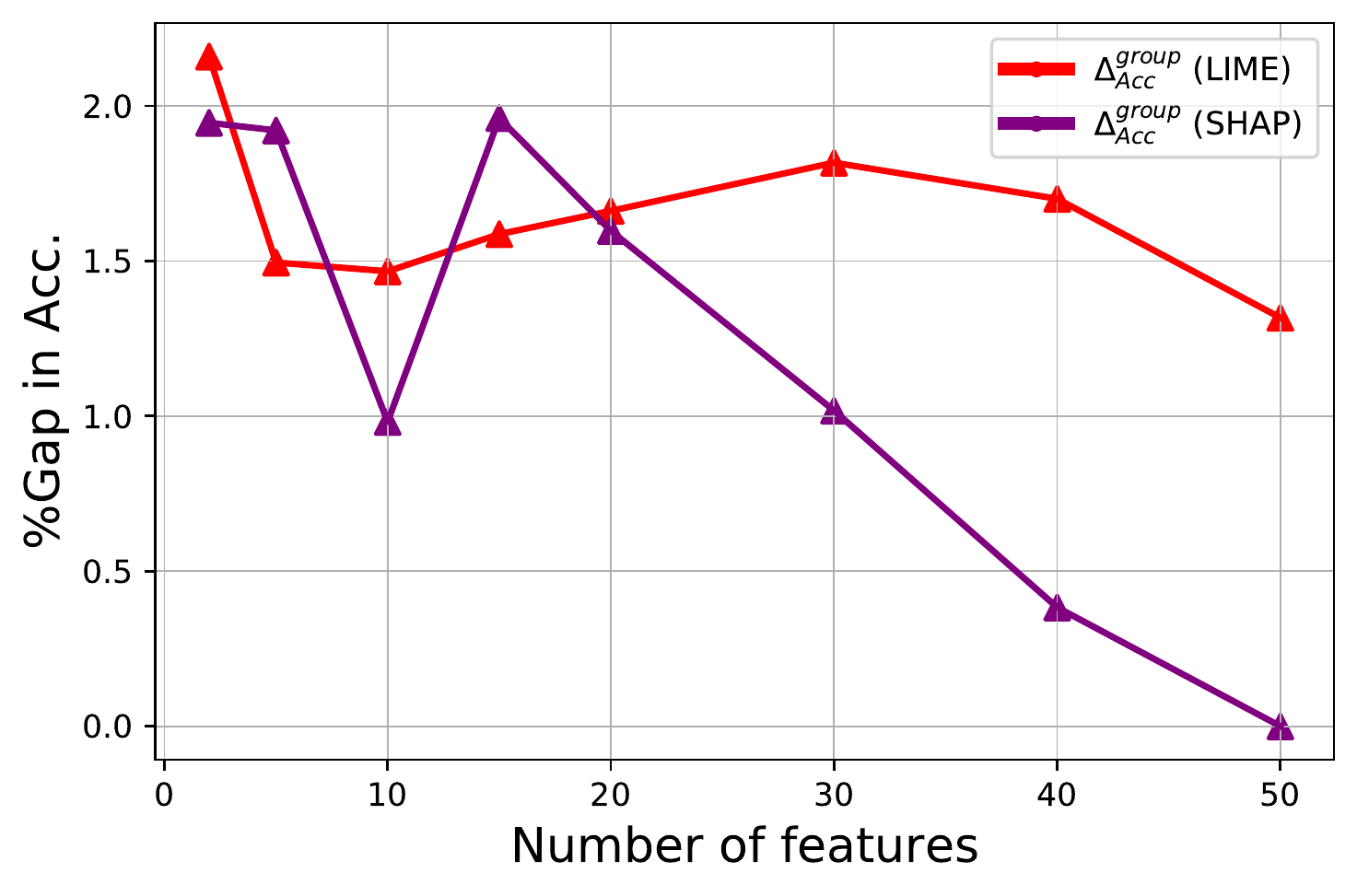}
         \caption{\texttt{mimic} + NN + Acc.-fidelity gap for local explanations}
    \end{subfigure}
    \begin{subfigure}{.49\textwidth}
        \includegraphics[width=1\linewidth]{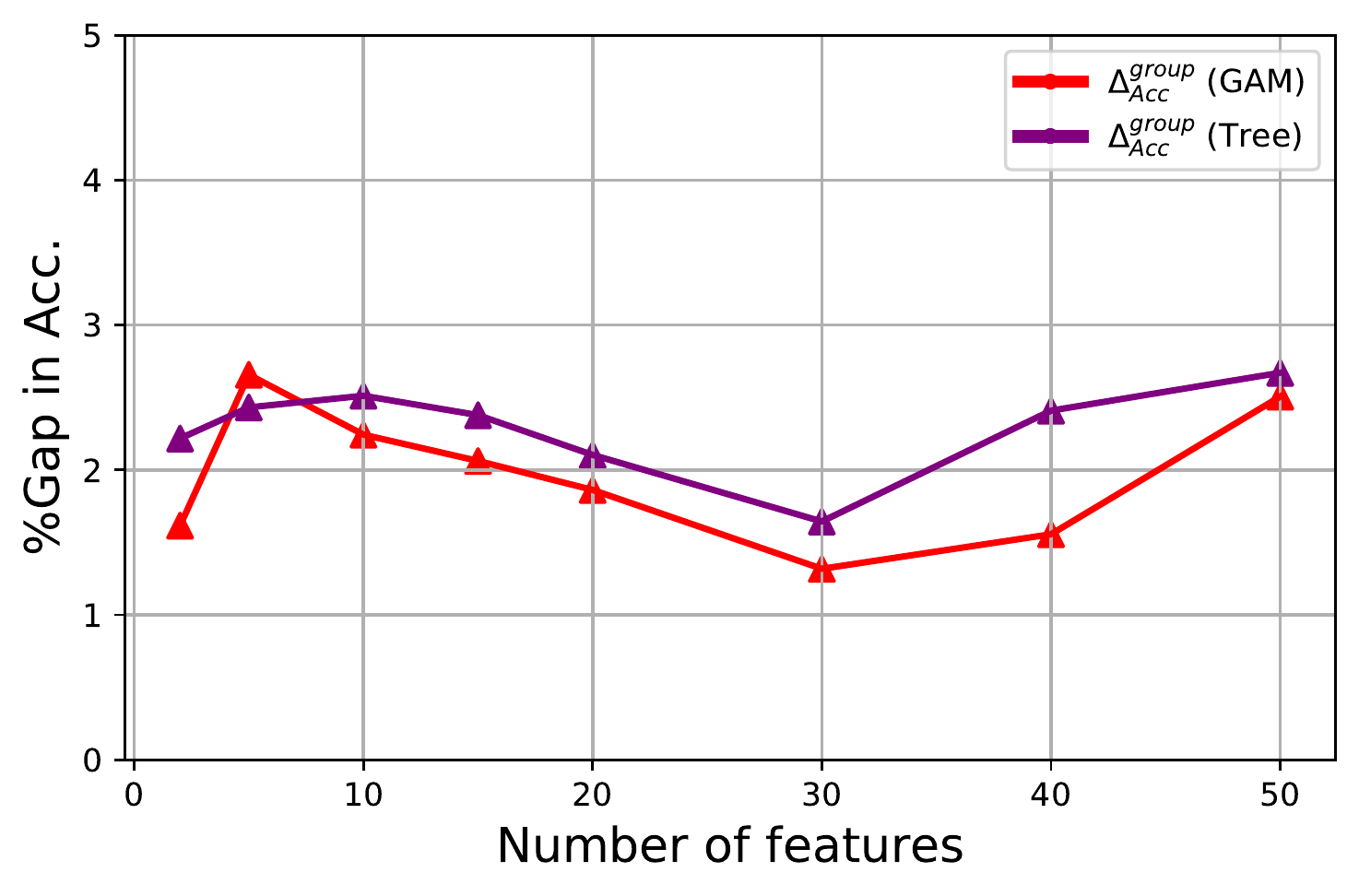}
         \caption{\texttt{mimic} + NN + Acc.-fidelity gap for global explanations}
    \end{subfigure}
    \caption{We often observe larger gaps (or larger mean approximation error)  with sparser models, i.e., fewer features in explanation model even in terms of accuracy-based fidelity. \label{fig:accuracy_gaps}}
\end{figure}

Similar to varying AUROC-fidelity gaps ($\Delta^{group}_{AUROC}$), the accuracy fidelity gaps also: (1) depend on number of features used in the explanation model, (2) are often higher when fewer features are used, especially in local explanation models (see Fig.\ref{fig:accuracy_gaps}). We also observe similar trends in the \texttt{adult} dataset (Fig.\ref{fig:auroc_gaps_adult}). 


\section{Worst-Case Fidelity}
We observe that the groups with worst-case fidelity varies depending on dataset, type of explanation model, and type of blackbox model. Generally, these groups are the minoritized (i.e., not the majority) subgroups based on a protected or demographic attribute in the dataset (see Table~\ref{tab:worst_case_groups}).
\begin{table*}[h]
\begin{tabular}{lllll}
\toprule
Dataset & Blackbox & Expl &      Worst-Case Group \\
\midrule
adult &             LR &               GAM &  Sex: Majority \\
 &             LR &              Tree &  Sex: Majority \\
 &             NN &               GAM &  Sex: Majority \\
 &             NN &              Tree &  Sex: Minority \\
lsac &             LR &               GAM &  Race: Majority \\
 &             LR &              Tree &  Race: Neither \\
 &             NN &               GAM &  Race: Neither \\
 &             NN &              Tree &  Race: Neither \\
mimic &             LR &               GAM &  Sex: Minority \\
 &             LR &              Tree &  Sex: Minority \\
 &             NN &               GAM &  Sex: Minority \\
 &             NN &              Tree &  Sex: Minority \\
recidivism &             LR &               GAM &  Race: Majority \\
 &             LR &              Tree &  Race: Minority \\
 &             NN &               GAM &  Race: Majority \\
 &             NN &              Tree &  Race: Majority \\\bottomrule
\end{tabular}
\caption{The subgroup with lowest fidelity for all global explanation models. The Worst-Case Group is in the form of `Protected Group: Majority/Minority/Neither Majority nor Minority` \label{tab:worst_case_groups}}
\end{table*}

\section{Effect of Varying Sampling Variance for LIME}
\begin{figure*}[h]
\centering
\begin{subfigure}{0.4\textwidth}
  \includegraphics[width=\textwidth]{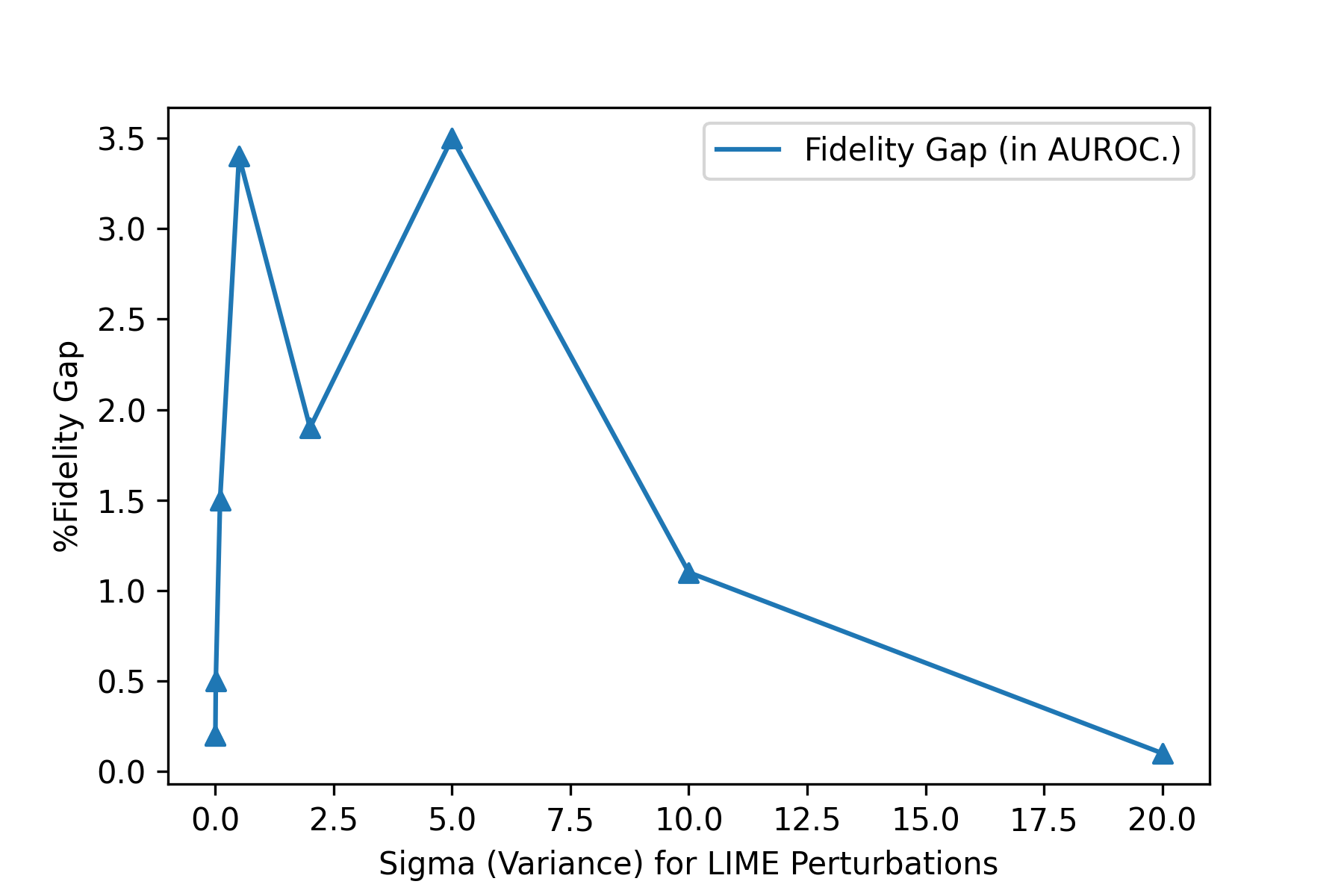}
\caption{Perturbation variance $\sigma$ vs $\Delta^{group}_{AUROC}$}
\end{subfigure} 
\begin{subfigure}{0.4\textwidth}
  \includegraphics[width=\textwidth]{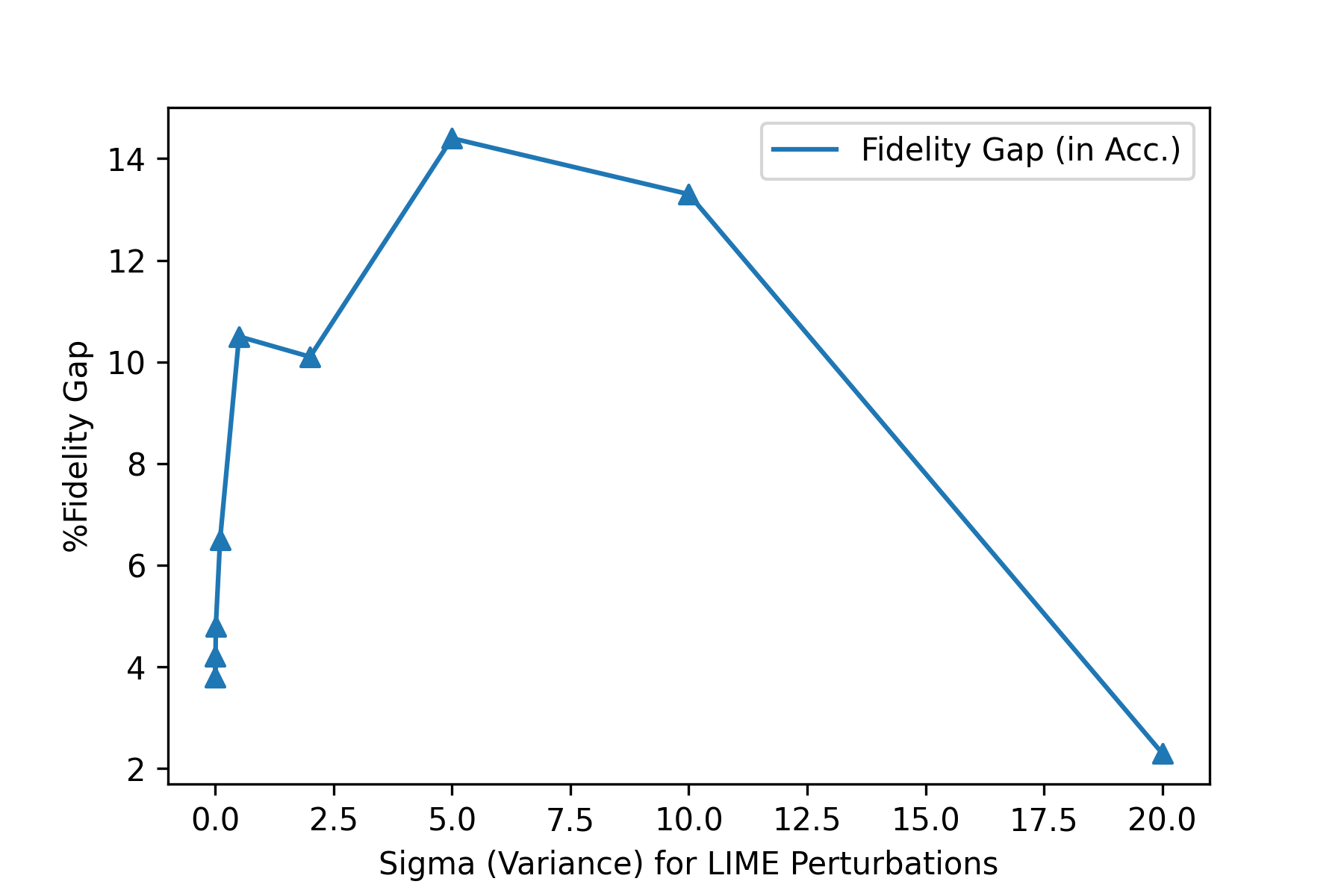}
\caption{Perturbation variance $\sigma$ vs $\Delta^{group}_{Acc.}$}
\end{subfigure}\\
    \caption{Fidelity gaps in AUROC and accuracy persist across a range of sampling variances}
    \label{fig:sigma_var_lime}
\end{figure*}

We observe that non-zero fidelity gaps in AUROC and accuracy persist across a range of sampling variances in LIME -- i.e., when $\sigma$ varies from 1e-3 to 20 as seen in Fig.~\ref{fig:sigma_var_lime}. The behaviour is as expected with low fidelity gaps for very low variance (sampling around very close to the test instance) and high variance (sampling from the whole distribution; not necessarily local). This makes sense, as LIME employs a weighted ridge regression, where sample errors are weighed inversely in proportion to their distance from test instance. When sampling variance is high, a lot of sampled data can lie far away (in constrast to standard LIME settings of ``local" neighborhood point with variance 1). And since weighting is inversely proportional to distance in loss function, the prediction for the test instance is always correct (we observe that fidelity is close to 100\% for all datapoints at high $\sigma$). When $\sigma$ is low, sample weights for all points are very high. If the blackbox model is stable to perturbations around points, then the fidelity should be high as well. However, if blackbox predictions fluctuate a lot with small perturbations to inputs, then fidelity gaps might exist (especially in Accuracy which is a threshold-based measure).

\section{Comparing Formulations of Empirical Risk Minimization and Just Train Twice}
\paragraph{Empirical risk minimization (ERM)}
 Empirical risk minimization minimizes the loss averaged across training points~\cite{liu2021just}. Following formulation from \citet{liu2021just}, given a loss function 
 $\ell(x, y; \theta) : \sX \times \sY \times \Theta \to \mathbb{R}$ (\textit{e.g.} cross-entropy loss), ERM minimizes the following objective: 
\begin{align}\label{eqn:erm}
    J_{\text{ERM}}(\theta) = \frac{1}{n} \sum \limits_{i=1}^n \ell(x_i, y_i; \theta).
\end{align}

LIME fits a locally accurate model fit on sampled synthetic datapoints around a test instance~\cite{ribeiro2016model} with weighted risk-minimization. However, since synthetic data is used in LIME, the group information of these synthetic instances is not known. Hence, the goal is to achieve good worst-group error at test time \emph{without training group annotations}. We adopt a recent method proposed for rebalancing and training robust models without group information: Just Train Twice (JTT)~\cite{liu2021just}. While LIME is generally trained by minimizing the weighted mean-squared-error between blackbox model and local linear model predictions with Empirical Risk Minimization (ERM), JTT relies on robust training that upsamples the error-prone or difficult regions to predict during optimization.

\paragraph{Just Train Twice (JTT)} 
JTT~\cite{liu2021just} is a two-stage approach that does not require group annotations at training time. First, it trains an identification model $\hat{f}_{\text{id}}$ via ERM and then identifies an error set $E = \{(x_i, y_i)~\text{s.t.}~ \hat{f}_{\text{id}}(x_i) \neq y_i\}$ of training examples that $\hat{f}_{\text{id}}$ misclassifies.
Then, it trains a final model $\hat{f}_{\text{final}}$ by upsampling the points in the identified error set~\cite{liu2021just}. 
\begin{align}
     J_{\text{up-ERM}}(\theta, E) = \Bigg( \upweightfactor \sum \limits_{(x, y) \in E} \!\!\ell(x, y; \theta) +  \sum \limits_{(x, y) \not\in E} \!\!\ell(x, y; \theta) \Bigg),
\end{align}
To practically implement this we LIME, we multiply the local distance-based weights with $\upweightfactor$ for training samples in the error set. We also threshold the outputs of $\hat{f}_{\text{id}}$ and $y_i$ to binary values for identifying the error set.

\section{Impact of Robust Training}
\begin{figure*}
\centering
\begin{subfigure}{0.49\textwidth}
  \includegraphics[width=\textwidth]{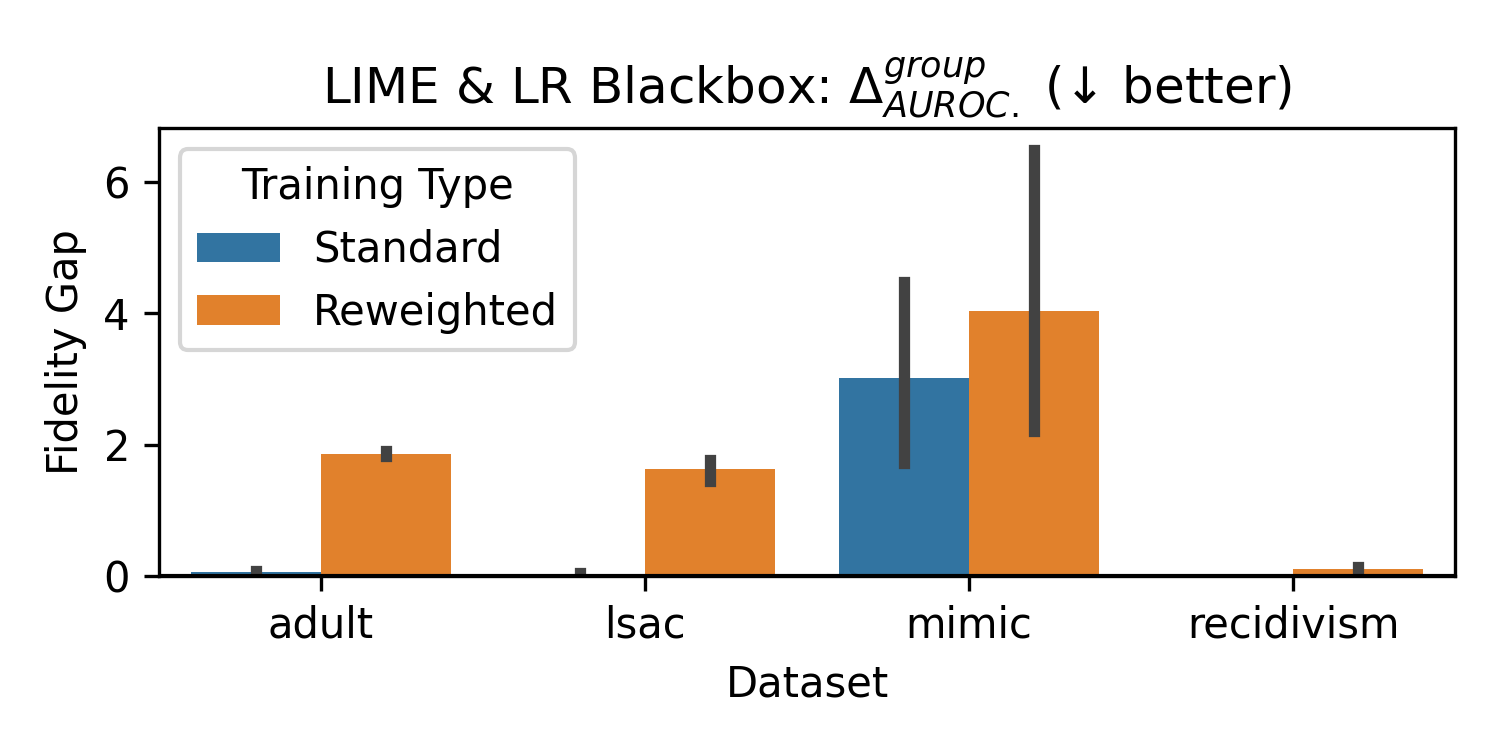}
\end{subfigure} 
\begin{subfigure}{0.49\textwidth}
  \includegraphics[width=\textwidth]{images/jtt_training_lime_nn_.png}
\end{subfigure}\\
\begin{subfigure}{0.49\textwidth}
  \includegraphics[width=\textwidth]{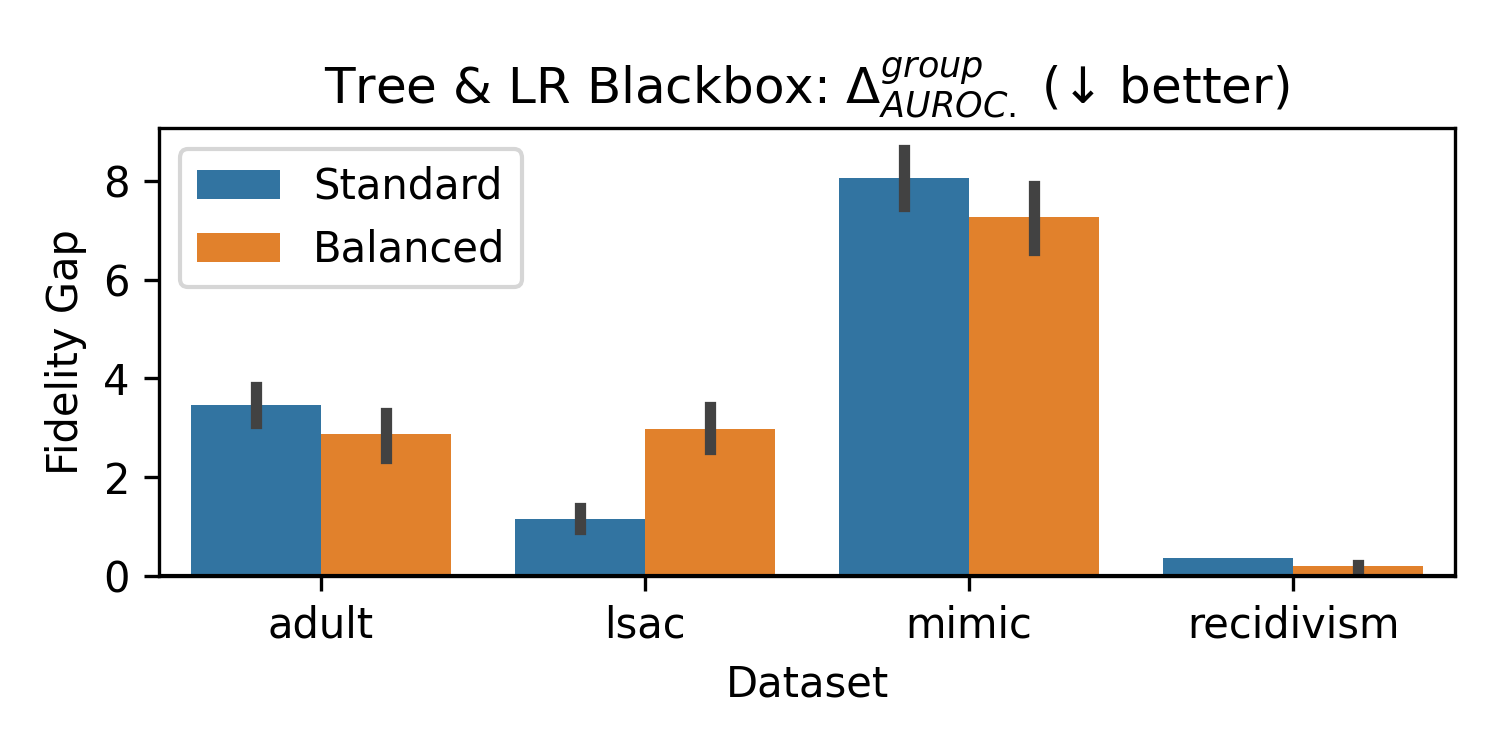}
\end{subfigure}
\begin{subfigure}{0.49\textwidth}
  \includegraphics[width=\textwidth]{images/balanced_nn.png}
\end{subfigure}
  \caption{AUROC Fidelity gaps across subgroups with and without robust training for LIME and Tree-based Models. We observe some improvements with balanced and robust training, but results depend on dataset and/or model type. Error bars indicate 95\% confidence intervals around mean fidelity gap in each case.}
  \label{fig:mod_gaps_lime_all}
\end{figure*}

In Fig.~\ref{fig:mod_gaps_lime_all}, we show the impact of robust and balanced training on all four datasets, two blackbox models, and two explanation (one local, one global) models.

\section{Simulation of Real-World Impact}
\label{app:sim}

We assume the following simulation parameters for whether the user makes a correct decision based on the correctness of the blackbox model and the fidelity explanation. 
\begin{itemize}
    \item $P(\text{User} = \text{Correct}\ |\ \text{BlackBox} = \text{Incorrect}, \text{Explanation} = \text{Good}) = 0.6497$. This was selected based on results from \citet{bansal2021does}.
    \item $P(\text{User} = \text{Correct}\ |\ \text{BlackBox} = \text{Incorrect}, \text{Explanation} = \text{Poor}) = 0.68$, as it is reasonable that a poor explanation would more likely reveal the error in the blackbox model to the user, resulting in greater likelihood of a correct decision due to lower trust in the model~\cite{papenmeier2019model}.
    
    \item $P(\text{User} = \text{Correct}\ |\ \text{BlackBox} = \text{Correct}, \text{Explanation} = \text{Good}) = 0.9281$. This was selected based on results from \citet{bansal2021does}.
    \item $P(\text{User} = \text{Correct}\ |\ \text{BlackBox} = \text{Correct}, \text{Explanation} = \text{Poor}) = 0.90$, as it is reasonable that a poor explanation could lead the user astray into believing that the blackbox was incorrect, leading to an incorrect decision~\cite{bansal2021does}.
\end{itemize}

Parameters specified above were obtained from the user study conducted by \citet{bansal2021does} for LIME explanations with the \emph{beer} sentiment review dataset (aggregated across question-participant data points). Note that these are \emph{subgroup-agnostic parameters}. While we use these parameters as reasonable estimates of real-world performance for our simulation, a detailed user-study is required to assess if these are accurate estimates and assumptions specifically for our case: this is a limitation of our simulation setup. Another assumption inherent in our simulation is that low fidelity explanations are poor. Thus our results are strongly influenced by the assumptions involved, but we believe still hold value as a proof-of-concept.

\begin{figure}[h]
    \centering
    \begin{subfigure}{0.5\textwidth}
          \centering
          \includegraphics[scale=0.5]{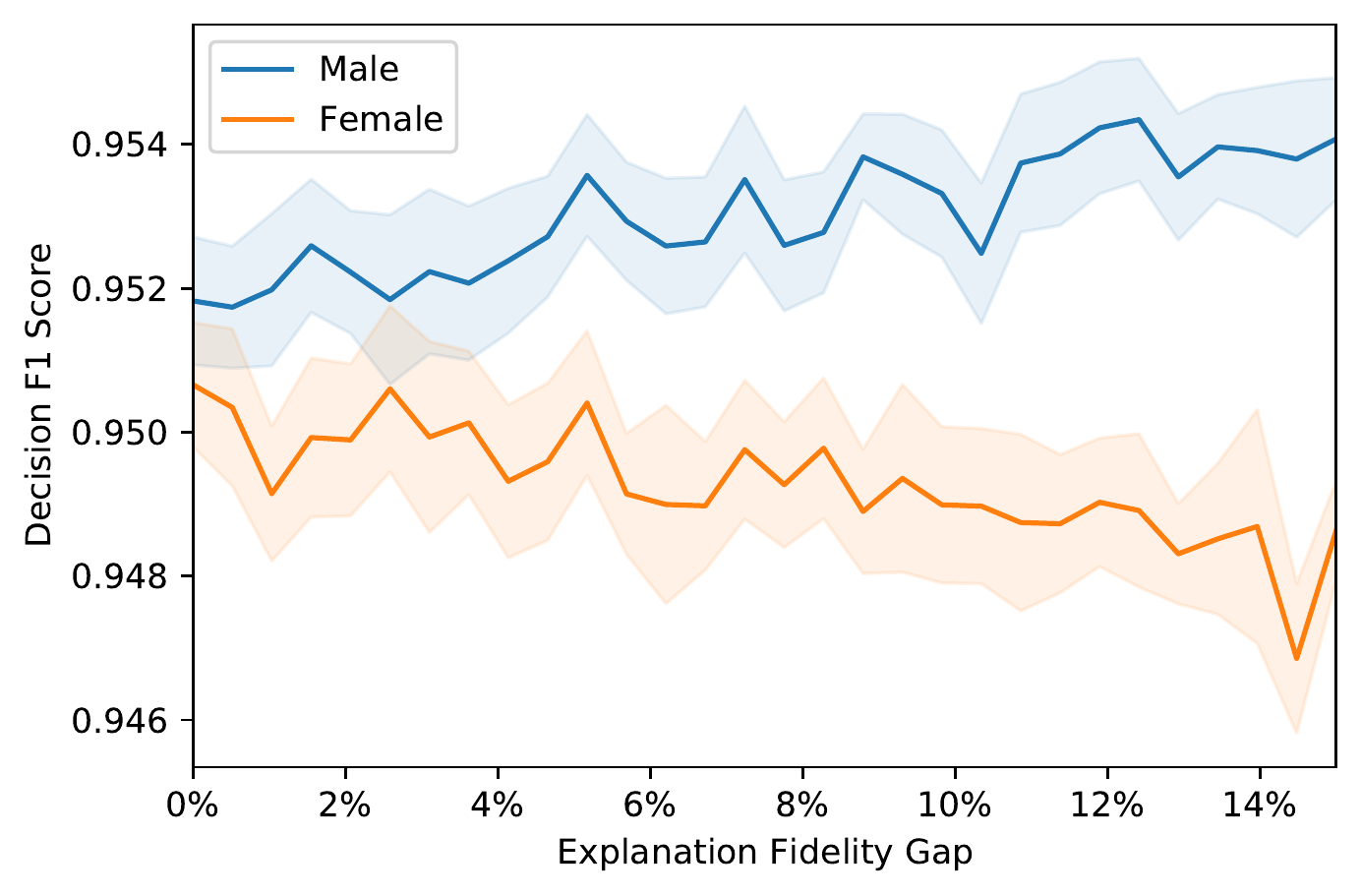}
          \caption{Logistic Regression}
          \label{fig:f1_expl_gap_sim_lr}
    \end{subfigure}%
    \begin{subfigure}{0.5\textwidth}
          \centering
          \includegraphics[scale=0.5]{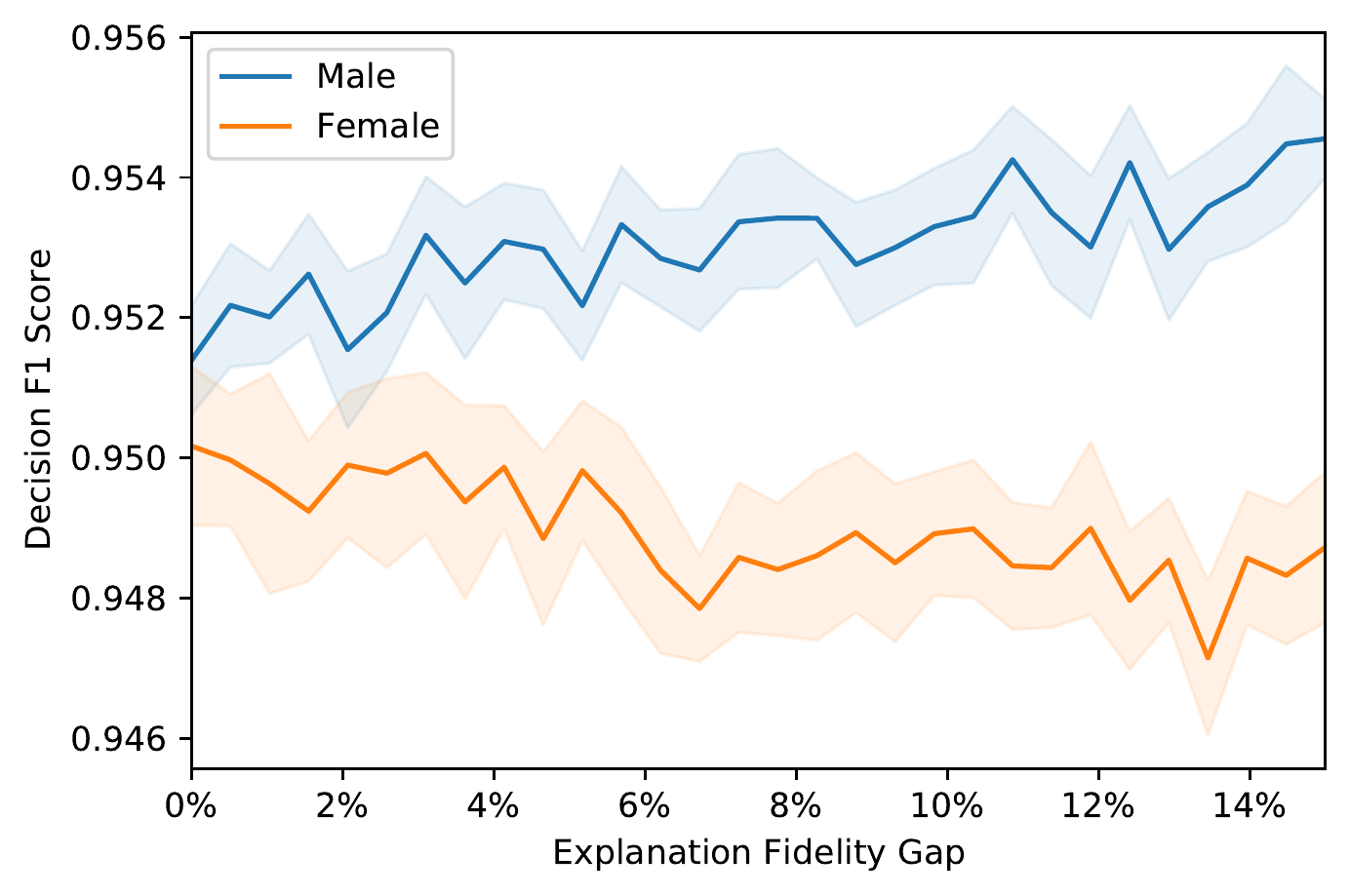}
          \caption{Neural Network}
          \label{fig:f1_expl_gap_sim_nn}
    \end{subfigure}%
    \caption{F1-score of an admission officer's decision versus the fidelity gap of the explanation method between males and females using (a) logistic regression and (b) neural network as the blackbox model. We simulate the real-world impact of a deployed blackbox machine learning model along with an explanation method with varying levels of fidelity between males and females. We find that larger gap in explanation fidelity leads to a larger decision accuracy gap between the groups. Error bars are 95\% confidence intervals derived from 20 runs of the simulation.}
    \label{fig:expl_gap_sim}
\end{figure}

For the blackbox model, we use the logistic regression and neural network models described in Section \ref{sec:experimental_setup}. For the explanation model, we assume that the average fidelity between males and females is 85\%, and vary the maximum fidelity gap between the mean and two groups from 0\% to 15\%. We report results with final decision-making accuracy in the main text, and also obtained similar results on computing the F1-score instead (see Fig.~\ref{fig:f1_expl_gap_sim_lr} and ~\ref{fig:f1_expl_gap_sim_nn}). The results shown here are all for blackbox models trained on a random 50\% \texttt{lsac} data split without label balancing via oversampling, tested on the remaining 50\%. Note that very similar results are obtained for models trained with balanced class labels via oversampling. 

\FloatBarrier


\section{Fair Blackbox Model}
We use an adversarial debiasing approach~\cite{zhang2018mitigating} to train fairer neural network blackbox models for two datasets: \texttt{lsac} and \texttt{mimic}. Optimal parameters are chosen based on lowest demographic fairness gap during validation. Note that we do not oversample datasets to deal with class imbalance, and rely on standard fair training approaches. Test performance is shown in Table~\ref{tab:fair_blackbox_groundtruth}, where we observe that models are fair(er).
\begin{table*}[h]
\begin{tabular}{llllrrrrr}
\toprule
Dataset & Blackbox  &  AUROC & Acc. & $\Delta_{AUROC}$ & $\Delta_{Acc.}$ & $\Delta^{group}_{AUROC}$ & $\Delta^{group}_{Acc.}$ & $|DP|$ \\
\midrule
lsac &     nn &         62.4\% ± 3.2\% &  91.8\% ± 1.4\% &   8.0\% ± 1.3\% &     10.3\% ± 2.1\% &   20.4\% ± 1.0\% &       6.0\% ± 0.7\% &          9.0\% ± 1.3\%  \\
mimic &             nn & 81.3\% ± 0.3\% &  89.2\% ± 0.3\% &   1.0\% ± 0.4\% &      0.9\% ± 0.3\% &    1.8\% ± 0.7\% &       1.6\% ± 0.5\% &    0.6\% ± 0.2\%\\
\bottomrule
\end{tabular}
\caption{Fair Neural-Network Blackbox Model Performance in Predicting Groundtruth\label{tab:fair_blackbox_groundtruth}}
\end{table*} 

\section{Removing Features Predictive of Protected Attribute}
We observe in Figure~4 that protected group information can be predicted from feature representations. Note that for datasets with more than two protected subgroups, we display the highest detection AUROC across all subgroups.   

On removing predictive features using methodology described in the main text, the results for \texttt{mimic} are shown in Table~\ref{tab:group_removed_subset}. The fidelity gaps using accuracy are not significantly higher than zero based on a one-sided Wilcoxon signed-rank test ($p>0.05$) for the logistic regression blackbox model, and are all less then $1$. These values are also reduced in comparison to the accuracy-based fidelity gaps on using all features. Blackbox and explanation models trained on similar versions of \texttt{lsac} dataset produce single class predictions due to high correlations between protected attributes and class label. We note that however that training models on these data representations lead to highly imbalanced blackbox model predictions (3\% positive class prevalence for \texttt{mimic}). We observed that oversampling the minority class during blackbox model training to mitigate the effects of this imbalance led to similar results for the \texttt{mimic} dataset (i.e., significant reduction in accuracy-based fidelity metrics).

\begin{table*}[h]
\begin{tabular}{llcccc}
        \toprule
        Blackbox & Expl  & $Fidelity^{AUROC}$ & $\Delta_{Acc.}$ & $\Delta^\text{group}_{AUROC}$ & $\Delta^\text{group}_{Acc.}$ 
         \\
        \midrule
         LR & GAM &  100.0\% ± 0.0\% &      0.0\% ± 0.0\% &    0.0\% ± 0.0\% &       0.0\% ± 0.0\% \\
        & Tree &  77.2\% ± 0.0\% &      0.0\% ± 0.0\% &    6.6\% ± 0.0\% &       0.1\% ± 0.0\% \\
        
        \midrule                     
        NN & GAM &  97.0\% ± 0.8\% &      0.2\% ± 0.2\% &    1.9\% ± 1.3\% &       0.5\% ± 0.3\% \\
        & Tree &  74.2\% ± 7.4\% &      0.2\% ± 0.3\% &    6.1\% ± 3.2\% &       0.6\% ± 0.2\% \\
\bottomrule
\end{tabular}
\caption{Fidelity gaps on \texttt{mimic} dataset when all features predictive of protected attribute label are removed \label{tab:group_removed_subset}}

\end{table*}

For the \texttt{lsac} dataset, we oversample the minority class label (i.e., data subgroups still imbalanced) during blackbox model training with the reduced representation. This helps avoid the single class prediction problem. Similar to the results for \texttt{mimic}, we find that fidelity gaps are significantly reduced (see Table~\ref{tab:group_removed_subset_lsac}). 

\begin{table*}[h]
\begin{tabular}{llccc}
        \toprule
        Blackbox & Expl  & $Fidelity^{AUROC}$ & $\Delta^\text{group}_{AUROC}$ & $\Delta^\text{group}_{Acc.}$ 
         \\
        \midrule
        NN & Tree &  100.0\% ± 0.0\% &    0.0\% ± 0.0\% &       0.0\% ± 0.0\% \\
        & GAM &  99.8\% ± 0.1\% &    0.6\% ± 0.5\% &       0.8\% ± 0.6\% \\
\bottomrule
\end{tabular}
\caption{Fidelity gaps on \texttt{lsac} dataset when all features predictive of protected attribute label are removed \label{tab:group_removed_subset_lsac}}

\end{table*}

However, the overall AUROC of the underlying NN blackbox classifier is low (0.62) with this representation. This indicates that class imbalance -- and varying degrees of class imbalance for data subgroups -- may be an important factor to consider. Our findings indicate that fidelity gaps persist across a range of class-imbalance ratios, but we leave the estimation of the effect of varying degrees of class imbalance (or positive-class prevalence) across subgroups with varying sample sizes on explanation fairness for future work. 

\end{document}